\RequirePackage{fix-cm}
\documentclass[10pt,a4paper,logo]{paper}

\usepackage[all]{hypcap}
\usepackage{fontawesome5}

\usepackage[authoryear, round]{natbib}

\usepackage{nicefrac}       %
\usepackage{multirow}
\usepackage[capitalize,nameinlink]{cleveref}
\usepackage{dsfont}
\usepackage{makecell}
\usepackage{subcaption}
\usepackage{bbm}
\usepackage{wrapfig}
\usepackage{mathtools,xparse}
\usepackage{array}
\usepackage{arydshln}
\usepackage{scalerel}
\usepackage{tablefootnote}
\usepackage{tocloft}
\usepackage{adjustbox}
\usepackage{setspace}
\usepackage[most,skins,theorems]{tcolorbox}

\newtcolorbox{promptbox}{
  colback=gray!5,
  colframe=gray!70,
  boxrule=0.8pt,
  arc=2mm,
  left=10pt,
  right=10pt,
  top=10pt,
  bottom=10pt,
  width=\linewidth,
  fontupper=\ttfamily\footnotesize,
  enhanced,
}

\newboolean{showsection}
\setboolean{showsection}{true}
\makeatletter
\@namedef{ver@everyshi.sty}{}
\makeatother

\definecolor{custom_green}{rgb}{0.0, 0.5, 0.0}
\definecolor{custom_red}{rgb}{1.0, 0.01, 0.24}
\definecolor{custom_blue}{HTML}{C9DAF7}
\definecolor{custom_purple}{HTML}{D9D1E9}
\definecolor{title_blue}{HTML}{204899}
\definecolor{cite_blue}{HTML}{044dc1}
\definecolor{cite_purple}{HTML}{7406a7}

\hypersetup{
    colorlinks = true,
    citecolor = {cite_blue},
    linkcolor = {cite_purple},
    urlcolor = {cite_purple},
}

\usepackage{amsmath,amsfonts,amssymb,bm,bbm,pifont,mathtools,mathdots,nicefrac,thmtools,thm-restate}

\def\bigset#1{\left\{#1\right\}}  %

\let\bar\overline

\def\1{\bm{1}}

\def\vh{{\bm{h}}}

\def\vu{{\bm{u}}}
\def\vv{{\bm{v}}}

\def\vx{{\bm{x}}}

\def\mB{{\bm{B}}}
\def\mC{{\bm{C}}}

\def\mE{{\bm{E}}}

\def\mU{{\bm{U}}}
\def\mV{{\bm{V}}}
\def\mW{{\bm{W}}}

\def\mSigma{{\bm{\Sigma}}}

\DeclareMathAlphabet{\mathsfit}{\encodingdefault}{\sfdefault}{m}{sl}
\SetMathAlphabet{\mathsfit}{bold}{\encodingdefault}{\sfdefault}{bx}{n}

\usepackage{type1cm}
\usepackage{float}
\usepackage{placeins}
\usepackage[normalem]{ulem}
\usepackage{enumerate}
\usepackage{tikz}
\usetikzlibrary{arrows.meta}
\usepackage{pgfplots}
\pgfplotsset{compat=1.18}
\usepgfplotslibrary{groupplots,fillbetween}

\makeatletter
\def\mathcolor#1#{\@mathcolor{#1}}
\def\@mathcolor#1#2#3{%
  \protect\leavevmode
  \begingroup
    \color#1{#2}#3%
  \endgroup
}
\makeatother

\Crefformat{equation}{#2Eq.\;(#1)#3}
\Crefformat{figure}{#2Figure #1#3}
\Crefformat{assumption}{#2Assumption #1#3}
\Crefname{assumption}{Assumption}{Assumptions}

\usepackage{crossreftools}
\makeatletter
\renewcommand\footnoterule{%
  \kern 15\p@
  \hrule \@width 2in \kern 2.6\p@
  \vspace{4pt}
}
\makeatother

\crefname{section}{Sec.}{Secs.}
\crefname{appendix}{Appx.}{Appcs.}
\crefname{equation}{Eq.}{Eqs.}
\crefname{definition}{Def.}{Defs.}
\crefname{figure}{Fig.}{Figs.}
\crefname{tabular}{Tab.}{Tabs.}
\crefname{table}{Tab.}{Tabs.}
\crefname{algorithm}{Alg.}{Algs.}
\crefname{theorem}{Thm.}{Thms.}
\crefname{lemma}{Lem.}{Lems.}
\crefname{proposition}{Prop.}{Props.}
\crefname{corollary}{Cor.}{Cors.}
\crefname{assumption}{Assum.}{Assums.}
\crefname{remark}{Rmk.}{Rmks.}

\newcommand{\method}[1]{\textsc{#1}}
\newcommand{\ESCal}{\method{ESCal}}
\newcommand{\Bridge}{\method{Bridge}}
\newcommand{\Surgery}{\method{Surgery}} 
\newcommand{\IntervMerge}{\method{IntervMerge}}
\newcommand{\FeatCal}{\method{FeatCal}}
\newcommand{\TA}{\method{Task Arithmetic}}
\newcommand{\AdaMerging}{\method{AdaMerging}}
\newcommand{\ProDistill}{\method{ProDistill}}
\newcommand{\TIES}{\method{TIES-Merging}}
\newcommand{\WUDI}{\method{WUDI-Merging}}

\title{Error-Subspace Calibration for Multi-Task Model Merging}

\reportnumber{}

\author{Yeseul Cho}
\author{Hanseul Cho}
\author[1]{Chulhee Yun}

\affil[1]{Kim Jaechul Graduate School of AI, KAIST}

\correspondingauthor{Correspondence to: \email{cyseul@kaist.ac.kr} and \email{chulhee.yun@kaist.ac.kr}.}

\begin{abstract}
Model merging combines task-specific expert models into a single model without joint training on their original datasets, but often incurs a performance gap relative to the individual experts.
Post-merging calibration can reduce this gap through task-specific representation alignment, but the alignment objective alone does not specify which feature directions to correct.
We propose \textit{Error-Subspace Calibration} (\textbf{\ESCal}), a post-merging calibration framework that identifies low-dimensional error subspaces from weight differences between expert and merged models.
At each intermediate block boundary, a \Bridge{} module projects the
merged representation onto the next block's input-projection error subspace and maps these coordinates to an additive correction in the current block's output-projection error subspace.
The subspace bases remain fixed, so calibration learns only these linear maps, yielding a parameter-efficient mechanism for feature correction.
We find that \ESCal{} brings the merged model closer to the experts in both representations and predictions, indicating more faithful recovery of task-specific behavior.
Experiments on vision and natural language understanding benchmarks show that \ESCal{} improves merged-model performance across diverse architectures and merging methods using a small calibration set for each task.
Our code is available at \href{https://github.com/cyseul/ESCal}{\faGithub \texttt{ESCal}}.
\end{abstract}

\begin{document}
\maketitle

\section{Introduction}
\label{sec:introduction}

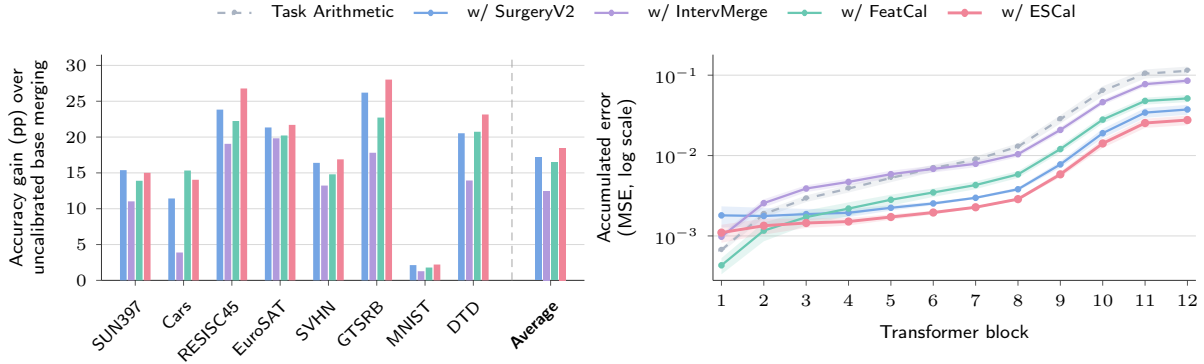
\begin{figure}[h]
    \centering
    \begingroup
\pgfplotstableread[col sep=comma]{Figures/derived_a_taskwise_improvement_pp.csv}\figOneTaskData
\pgfplotstablegetrowsof{\figOneTaskData}
\edef\figOneTaskCount{\pgfplotsretval}
\newcommand{\figOneComputeMean}[2]{%
  \pgfmathsetmacro{#2}{0}%
  \pgfplotstableforeachcolumnelement{#1}\of\figOneTaskData\as\figOneCell{%
    \pgfmathsetmacro{#2}{#2+\figOneCell/\figOneTaskCount}%
  }%
}
\figOneComputeMean{SurgeryV2}{\figOneSurgeryMean}
\figOneComputeMean{FeatCal}{\figOneFeatCalMean}
\figOneComputeMean{ESCal}{\figOneESCalMean}

\pgfmathsetmacro{\figOneIntervMean}{12.5065263327}

\definecolor{figOneInterv}{HTML}{B19ADC}
\definecolor{figOneSurgery}{HTML}{75A5E4}
\definecolor{figOneFeatCal}{HTML}{69C8B2}
\definecolor{figOneESCal}{HTML}{F08598}
\definecolor{figOneTA}{HTML}{A9B3C2}
\pgfplotsset{
  fig one axis/.style={
    width=0.48\linewidth,
    height=0.275\linewidth,
    font=\sffamily\fontsize{7}{8}\selectfont,
    tick label style={font=\sffamily\fontsize{6.5}{7.5}\selectfont},
    label style={font=\sffamily\fontsize{7}{8}\selectfont},
    axis line style={black!80,line width=0.35pt},
    tick style={black!65,line width=0.3pt},
    tick align=outside,
    major tick length=2pt,
    axis x line*=bottom,
    axis y line*=left,
    ymajorgrids=true,
    xmajorgrids=false,
    grid style={black!15,line width=0.25pt},
    legend style={
      at={(0.02,0.985)},anchor=north west,
      draw=none,fill=none,
      font=\sffamily\fontsize{6.5}{7.5}\selectfont,
      cells={anchor=west},
      inner xsep=1pt,inner ysep=1pt,
      row sep=-0.5pt,
    },
    legend image post style={xscale=0.75},
  },
  fig one curve/.style={line width=0.85pt,mark=*,mark size=0.8pt},
}
\begin{tikzpicture}
\begin{groupplot}[
    group style={group size=2 by 1,horizontal sep=17mm},
    fig one axis,
]
\nextgroupplot[
    ybar=0.5pt,
    bar width=2.5pt,
    xmin=-0.7,xmax=9.3,
    ymin=0,ymax=31.5,
    xtick={0,1,2,3,4,5,6,7,8.6},
    xticklabels={SUN397,Cars,RESISC45,EuroSAT,SVHN,GTSRB,MNIST,DTD,\textbf{Average}},
    x tick label style={rotate=45,anchor=north east},
    ytick={0,5,10,15,20,25,30},
    ylabel={Accuracy gain~(pp) over\\uncalibrated base merging},
    ylabel style={align=center},
]
\addplot[fill=figOneSurgery,fill opacity=1,draw=none,bar shift=-4.5pt]
  table[x expr=\coordindex,y=SurgeryV2,col sep=comma]
  {Figures/derived_a_taskwise_improvement_pp.csv};
\addplot[fill=figOneInterv,fill opacity=1,draw=none,bar shift=-1.5pt]
  table[x=x,y=gain,col sep=comma] {Figures/figure1-intervmerge-gain.csv};
\addplot[fill=figOneFeatCal,fill opacity=1,draw=none,bar shift=1.5pt]
  table[x expr=\coordindex,y=FeatCal,col sep=comma]
  {Figures/derived_a_taskwise_improvement_pp.csv};
\addplot[fill=figOneESCal,fill opacity=1,draw=none,bar shift=4.5pt]
  table[x expr=\coordindex,y=ESCal,col sep=comma]
  {Figures/derived_a_taskwise_improvement_pp.csv};

\draw[black!35,line width=0.35pt,densely dashed]
  (axis cs:7.8,0) -- (axis cs:7.8,31.5);
\addplot[fill=figOneSurgery,fill opacity=1,draw=none,bar shift=-4.5pt,forget plot]
  coordinates {(8.6,\figOneSurgeryMean)};
\addplot[fill=figOneInterv,fill opacity=1,draw=none,bar shift=-1.5pt,forget plot]
  coordinates {(8.6,\figOneIntervMean)};
\addplot[fill=figOneFeatCal,fill opacity=1,draw=none,bar shift=1.5pt,forget plot]
  coordinates {(8.6,\figOneFeatCalMean)};
\addplot[fill=figOneESCal,fill opacity=1,draw=none,bar shift=4.5pt,forget plot]
  coordinates {(8.6,\figOneESCalMean)};

\nextgroupplot[
    ymode=log,
    log basis y=10,
    xmin=0.8,xmax=12.2,
    ymin=0.00028,ymax=0.18,
    xtick={1,...,12},
    ytick={0.001,0.01,0.1},
    minor y tick num=8,
    yminorgrids=true,
    minor grid style={black!7,line width=0.2pt},
    xlabel={Transformer block},
    ylabel={Accumulated error\\(MSE, log scale)},
    ylabel style={align=center},
    legend to name=figOneSharedLegend,
    legend columns=5,
    legend style={column sep=5pt},
]
\addplot[draw=none,name path=figOneTAlow,forget plot]
  table[x=block,y=TA_lower,col sep=comma]{Figures/figure1-accumulated-error.csv};
\addplot[draw=none,name path=figOneTAhigh,forget plot]
  table[x=block,y=TA_upper,col sep=comma]{Figures/figure1-accumulated-error.csv};
\addplot[fill=figOneTA,fill opacity=0.17,draw=none,forget plot]
  fill between[of=figOneTAlow and figOneTAhigh];
\addplot[draw=none,name path=figOneSurgeryLow,forget plot]
  table[x=block,y=SurgeryV2_lower,col sep=comma]{Figures/figure1-accumulated-error.csv};
\addplot[draw=none,name path=figOneSurgeryHigh,forget plot]
  table[x=block,y=SurgeryV2_upper,col sep=comma]{Figures/figure1-accumulated-error.csv};
\addplot[fill=figOneSurgery,fill opacity=0.17,draw=none,forget plot]
  fill between[of=figOneSurgeryLow and figOneSurgeryHigh];
\addplot[draw=none,name path=figOneFeatLow,forget plot]
  table[x=block,y=FeatCal_lower,col sep=comma]{Figures/figure1-accumulated-error.csv};
\addplot[draw=none,name path=figOneFeatHigh,forget plot]
  table[x=block,y=FeatCal_upper,col sep=comma]{Figures/figure1-accumulated-error.csv};
\addplot[fill=figOneFeatCal,fill opacity=0.17,draw=none,forget plot]
  fill between[of=figOneFeatLow and figOneFeatHigh];
\addplot[draw=none,name path=figOneESCalLow,forget plot]
  table[x=block,y=ESCal_lower,col sep=comma]{Figures/figure1-accumulated-error.csv};
\addplot[draw=none,name path=figOneESCalHigh,forget plot]
  table[x=block,y=ESCal_upper,col sep=comma]{Figures/figure1-accumulated-error.csv};
\addplot[fill=figOneESCal,fill opacity=0.17,draw=none,forget plot]
  fill between[of=figOneESCalLow and figOneESCalHigh];
\addplot[draw=none,name path=figOneIntervLow,forget plot]
  table[x=block,y=lower,col sep=comma]{Figures/figure1-intervmerge-error.csv};
\addplot[draw=none,name path=figOneIntervHigh,forget plot]
  table[x=block,y=upper,col sep=comma]{Figures/figure1-intervmerge-error.csv};
\addplot[fill=figOneInterv,fill opacity=0.17,draw=none,forget plot]
  fill between[of=figOneIntervLow and figOneIntervHigh];

\addplot[fig one curve,color=figOneTA,dashed]
  table[x=block,y=TA,col sep=comma]{Figures/figure1-accumulated-error.csv};
\addlegendentry{Task Arithmetic}
\addplot[fig one curve,color=figOneSurgery]
  table[x=block,y=SurgeryV2,col sep=comma]{Figures/figure1-accumulated-error.csv};
\addlegendentry{w/ SurgeryV2}
\addplot[fig one curve,color=figOneInterv]
  table[x=block,y=mean,col sep=comma]{Figures/figure1-intervmerge-error.csv};
\addlegendentry{w/ IntervMerge}
\addplot[fig one curve,color=figOneFeatCal]
  table[x=block,y=FeatCal,col sep=comma]{Figures/figure1-accumulated-error.csv};
\addlegendentry{w/ FeatCal}
\addplot[fig one curve,color=figOneESCal,line width=1.1pt,mark size=1pt]
  table[x=block,y=ESCal,col sep=comma]{Figures/figure1-accumulated-error.csv};
\addlegendentry{w/ ESCal}
\end{groupplot}
\path (group c1r1.north west) -- (group c2r1.north east)
  node[midway,above=6pt]{\pgfplotslegendfromname{figOneSharedLegend}};
\end{tikzpicture}
\endgroup
    \caption{
        \textbf{Post-merging calibration of an eight-task \TA{} merge
        of ViT-B/32 experts.}
        \textbf{(Left)} Test-accuracy gains over the uncalibrated merge
        (percentage points); ``Average'' denotes the eight-task mean.
        \textbf{(Right)} Mean squared error between merged and expert
        hidden features at depth $\ell$, averaged over eight tasks;
        shading indicates standard errors across tasks.
        IntervMerge uses the selected ViT-B/32--Task Arithmetic setting ($r=16$, $\eta=3\times10^{-4}$, 1,000 outer iterations).
        Its results are averaged over five seeds in both panels; for feature error, seed numerators and element counts are pooled within each task before averaging across tasks.
        \ESCal{} achieves the largest accuracy gains on six tasks
        and on average, and the lowest feature error at most block
        boundaries.
    }
    \label{fig:analysis}
\end{figure}

Model merging aims to integrate the capabilities of multiple task-specific \textit{expert} models into a single \textit{merged} model without jointly training it on their original datasets~\citep{ilharco2023ta,yadav2023ties, cheng2025wudi,xu2025prodistill}.
By reusing fine-tuned expert models, it offers a practical alternative to multi-task training or deploying a separate model for each task.
However, the merged model often performs worse than the expert models, leaving a performance gap that limits the effectiveness of multi-task model merging.
Reducing this gap while retaining the efficiency of model merging
remains an important research question~\citep{yang2026ModelMergingSurvey}.

To narrow this gap, post-merging calibration methods use a small calibration set to align the representations of merged models with those of the corresponding experts~\citep{yang2024surgery,yang2024surgeryv2,gu2026featcal}.
The expert representations serve as alignment targets, enabling calibration without labels.
Calibration can be applied at the final representation or at intermediate layers to address discrepancies that arise throughout the network.

Although representation alignment is effective, the alignment objective alone does not explicitly specify the feature directions that calibration modules can modify. For example, \Surgery{} and \Surgery{V2} learn the projection matrices of their low-rank calibration modules from calibration samples, with the correction directions determined by the alignment objective. This motivates us to ask whether these directions can be identified from merging errors before calibration, leaving fewer parameters to learn from the calibration data.
\textit{Can we improve both the effectiveness and efficiency of representation alignment by restricting calibration to the feature subspaces associated with merging errors?}

To investigate this question, we examine \textit{merging error}, defined as the difference between expert weights and the merged weights.
Its dominant singular subspaces capture the largest components of these weight differences.
Based on this insight, we propose \textit{Error-Subspace Calibration} (\textbf{\ESCal}), which constructs task-specific calibration subspaces from merging errors.
At each intermediate block boundary, a \Bridge{} module reads coordinates from the next block's input-projection error subspace and maps them to an additive correction in the current block's output-projection error subspace. 
With both subspaces fixed, calibration learns only a mapping between them (Section~\ref{sec:method}).

We evaluate \ESCal{} on vision and natural language understanding benchmarks across various architectures and base merging methods, with additional experiments on natural language generation.
\ESCal{} improves merged-model performance using a small calibration set per task (Section~\ref{sec:experiments}).
Our ablation studies examine how subspace construction and constraints affect calibration performance (Section~\ref{sec:error_subspace_construction}).
Our alignment analyses further show that \ESCal{} brings the merged model closer to the expert models in both representations and predictions (Section~\ref{sec:prediction_alignment}).

Keeping the subspace bases fixed also makes \ESCal{} parameter-efficient, as each \Bridge{} module learns only an $r\times r$ mapping for each task and block boundary.
Across all evaluated configurations, \ESCal{} uses fewer trainable
parameters than \Surgery{V2}, while achieving higher average
performance in most settings (Section~\ref{sec:train_efficiency}).

\section{Related Work}

\subsection{Multi-task Model Merging}
A foundational approach is \TA{}~\citep{ilharco2023ta}, which represents the knowledge acquired during fine-tuning as a \emph{task vector}, defined as the difference between the fine-tuned and pretrained model parameters.
It constructs a multi-task model by adding the sum of task vectors, scaled by a single merging coefficient, to the pretrained model parameters.
Building on this approach, \AdaMerging{}~\citep{yang2024adamerging} adaptively learns task-wise or layer-wise merging coefficients using unlabeled test data, rather than relying on a single fixed coefficient.

A major line of work focuses on mitigating interference among task updates.
\TIES{}~\citep{yadav2023ties} reduces interference by removing redundant parameter updates and resolving sign conflicts before merging.
\WUDI{}~\citep{cheng2025wudi} leverages task-vector-induced subspaces to reduce parameter interference without additional data.
\ProDistill{}~\citep{xu2025prodistill} instead uses few-shot data to progressively calibrate intermediate features through layer-wise distillation with element-wise merging coefficients, allowing more fine-grained adaptation of the merged model toward task-specific experts.
These approaches modify how the merged model is constructed, in contrast to post-merging calibration methods that operate on an already merged checkpoint.
Additional related work on model merging is discussed in Appendix~\ref{app:additional_related_work}.

\subsection{Post-merging Calibration}
Post-merging calibration aims to reduce the performance gap between the merged model and its task-specific experts by correcting representation discrepancies after merging.

\Surgery{}~\citep{yang2024surgery} characterizes these discrepancies as
\emph{representation bias} and learns a lightweight task-specific
low-rank module to align the merged model's final-layer representation
with that of the corresponding expert.
Its projection matrices are learned from unlabeled test samples, allowing the
directions of the representation correction to be determined by the alignment objective.
\method{ProbSurgery}~\citep{wei2025probsurgery} extends this framework
by probabilistically modeling representation bias and its uncertainty.

While these approaches calibrate only the final representation,
\Surgery{V2}~\citep{yang2024surgeryv2} extends calibration to intermediate
layers, motivated by the observation that representation bias can arise
throughout the network and accumulate with depth.
It places task-specific calibration modules at multiple locations and aligns the merged representations with those of the corresponding expert.
\IntervMerge{}~\citep{osial2024interv} uses a similar layer-wise correction strategy, formulating task-specific feature adjustment as low-rank interventions within Transformer blocks.
Its parameter-efficient mini-interventions restrict the correction to selected dimensions of the intermediate representation.

More recently, \FeatCal{}~\citep{gu2026featcal} takes a different approach by
interpreting the discrepancy between merged and expert representations
as \emph{feature drift}.
It explicitly accounts for both locally introduced errors and their
propagation through subsequent layers and, rather than introducing
additional calibration modules, directly recalibrates the merged weights
layer by layer using closed-form updates.
In contrast, \ESCal{} keeps the merged weights frozen and learns only small mappings between fixed error subspaces derived from weight differences between the expert and merged models.

\begin{figure}[!t]
    \centering
        \resizebox{0.95\linewidth}{!}{\input{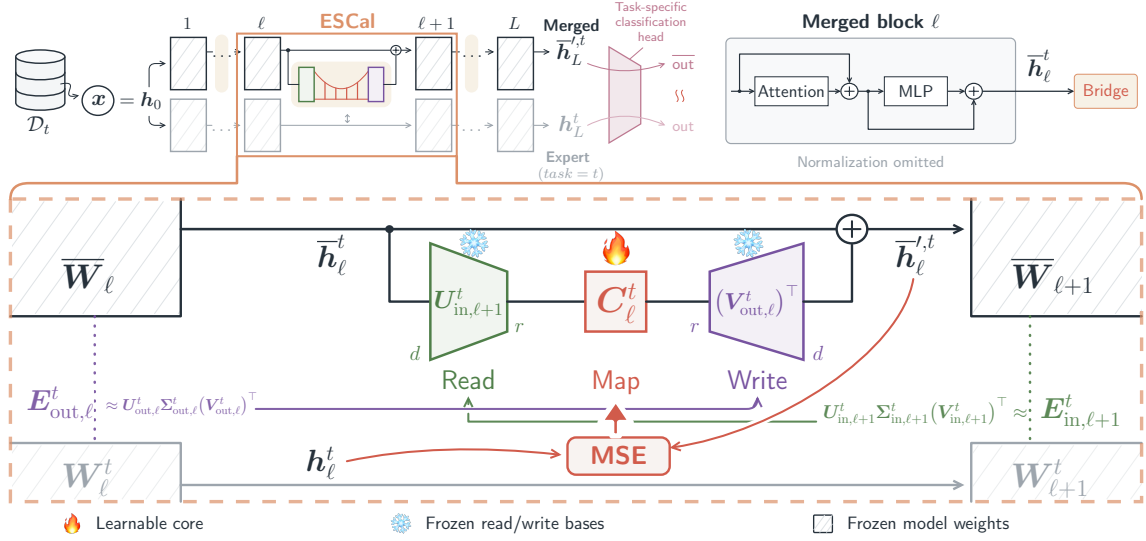}}
    \caption{
        \textbf{Overview of Error-Subspace Calibration (\ESCal).}
        Task-specific \Bridge{} modules align the frozen merged model
        with its expert at complete block outputs, after both residual
        additions.
        At intermediate boundaries, read and write bases come from
        merging errors in the next block's joint QKV input projection
        and the current block's MLP output projection, respectively.
    }
    \label{fig:escal_framework}
    \vspace{-6pt}
\end{figure}

\vspace{-8pt}
\section{Method: Error-Subspace Calibration}
\vspace{-2pt}
\label{sec:method}
We characterize the discrepancy introduced by model merging through low-dimensional error subspaces derived from weight differences between expert and merged models, and use them to constrain post-merging calibration.
The subspaces determine where a calibration module reads from and where its correction can act, leaving only the mapping between them to be learned.

\vspace{-2pt}
\paragraph{Notation.}
For a positive integer $N$, we write $[N]=\bigset{1,\ldots,N}$.
We consider $T$ tasks and a Transformer backbone with $L$ blocks,
using $t\in[T]$ to identify a task and $\ell\in[L]$ to identify a block.
For a given linear transformation in block $\ell$, $\mW^t_\ell$
denotes the weight matrix of the expert for task $t$, while
$\bar{\mW}_\ell$ denotes the corresponding weight matrix of the merged model.
The merged weights are shared across tasks so we omit the task index.
Throughout, we represent hidden features as row vectors: an input
$\vx$ is multiplied by the weight matrix on the right, giving
$\vx\mW$ for a linear transformation.

\subsection{Merging Error Subspaces}

For a linear transformation, the discrepancy between expert and merged outputs under a common input is determined by their weight difference.
We define the merging error as
\begin{equation}
    \mE_\ell^t = \mW_\ell^t - \bar{\mW}_\ell.
\end{equation}

Write its compact singular value decomposition (SVD) as
\begin{equation}
    \mE_\ell^t
    =
    \mU_\ell^t \mSigma_\ell^t
    \left(\mV_\ell^t\right)^\top,
\end{equation}
where $\mSigma_\ell^t$ contains all nonzero singular values in
descending order.

Under our row-vector convention, the columns of $\mU_\ell^t$
and $\mV_\ell^t$ span the input-side and output-side error
subspaces, respectively.
For a common input $\vx$, the output discrepancy is
$\vx\mE_\ell^t =
(\vx\mU_\ell^t)\mSigma_\ell^t(\mV_\ell^t)^\top$.
Thus, the discrepancy depends only on the input coordinates
$\vx\mU_\ell^t$, and its transpose lies in
$\operatorname{span}(\mV_\ell^t)$.

We retain the leading $r$ left and right singular vectors to
construct low-dimensional approximations of these subspaces.
For a fixed rank budget, truncated SVD minimizes the approximation
error of the weight difference in Frobenius norm, supporting
the choice of leading singular directions.
This argument concerns the local output discrepancy of a linear
transformation under a common input.
Appendix~\ref{app:error_subspace_motivation} provides the linear-map analysis underlying this argument.

\subsection{Cross-Layer Error-Subspace Calibration}
Feature drift in a merged model reflects both locally introduced
mismatch and drift propagated from preceding layers~\citep{gu2026featcal}.
Thus, a discrepancy at one block also affects the representations
processed by subsequent blocks.
This propagation view suggests calibrating at block
boundaries, where a representation serves as both the output of the
current block and the input to the next.

For a selected linear transformation under a common input, its weight-induced output discrepancy depends only on the input's coordinates in its full input-side error subspace.
This motivates using the next block's input-projection error subspace to select the coordinates the calibration module reads, while using the current block's output-projection error subspace to specify its correction directions.

\paragraph{Read and write bases.}
For each Transformer block \(\ell\), we construct the read basis from the leading input-side error directions of the joint query, key, and value (QKV) input projection, and the write basis from the leading output-side error directions of the MLP output projection. We refer to the resulting subspaces as the input-projection and output-projection error subspaces, respectively.
We denote the expert weights of these two projections by $\mW_{\mathrm{in},\ell}^t$ and
$\mW_{\mathrm{out},\ell}^t$.
We write both projection weights in input-by-output form.
The input projection has $d$ rows, and the output projection has $d$ columns, so their left and right singular vectors, respectively, lie in the same $d$-dimensional hidden representation space.

Applying the merging error definition to these projections gives
\begin{equation}
    \mE_{\mathrm{in},\ell}^t
    =
    \mW_{\mathrm{in},\ell}^t - \bar{\mW}_{\mathrm{in},\ell},
    \qquad
    \mE_{\mathrm{out},\ell}^t
    =
    \mW_{\mathrm{out},\ell}^t - \bar{\mW}_{\mathrm{out},\ell}.
\end{equation}
We use $\mU_{\mathrm{in},\ell}^t$ and $\mV_{\mathrm{out},\ell}^t$ to denote their top-$r$ left and right singular-vector bases, respectively.

At the boundary between blocks $\ell$ and $\ell+1$, we pair the read basis $\mU_{\mathrm{in},\ell+1}^t$ with the write basis
$\mV_{\mathrm{out},\ell}^t$.
This pairing reads coordinates along the dominant input-side error directions of block $\ell+1$'s QKV input projection and writes corrections along the dominant output-side error directions of block $\ell$'s MLP output projection.

\paragraph{\Bridge{} module.}
Using these fixed rank-$r$ bases, we define the task-specific
\Bridge{} module at each intermediate block boundary
$\ell=1,\ldots,L-1$ as
\begin{equation}
    \mB_\ell^t
    =
    \mU_{\mathrm{in},\ell+1}^t
    \mC_\ell^t
    \left(\mV_{\mathrm{out},\ell}^t\right)^\top,
    \qquad
    \mC_\ell^t \in \mathbb{R}^{r\times r},
    \label{eq:bridge}
\end{equation}
where the core $\mC_\ell^t$ maps coordinates in the read basis
to correction coefficients in the write basis.
Only $\mC_\ell^t$ is trainable, requiring $r^2$ parameters
per task and location.

At the terminal location, where there is no following Transformer block, we use the task-specific output head's top-$r$ input singular vectors as the read basis $\mU_{\mathrm{head}}^t$, giving
\begin{equation}
    \mB_L^t
    =
    \mU_{\mathrm{head}}^t
    \mC_L^t
    \left(\mV_{\mathrm{out},L}^t\right)^\top.
\end{equation}
Unlike the intermediate read bases, $\mU_{\mathrm{head}}^t$ is
derived from the head weights,
capturing directions to which the head is most sensitive.

\subsection{Module Training and Feature Calibration}
Calibration is performed independently for each task $t$ using a task-specific calibration set $\mathcal D_t$.
The read and write bases are computed once and kept fixed, while both the merged model and the corresponding expert remain frozen.
We optimize only the \Bridge{} cores $\mC_\ell^t$ to align the merged model's representations with those of the expert.

For an input $\vx$, let $\bar{\vh}_\ell^t(\vx)$ denote the output of merged Transformer block $\ell$ after both residual additions in its self-attention and MLP sublayers, including \Bridge{} corrections propagated from preceding block boundaries.
At each intermediate boundary $\ell=1,\ldots,L-1$, \Bridge{} applies the residual correction
\begin{equation}
    \begin{aligned}
    \bar{\vh}_\ell^{\prime,t}(\vx)
    &=
    \bar{\vh}_\ell^t(\vx)
    +
    \bar{\vh}_\ell^t(\vx) \mB_\ell^t \\
    &=
    \bar{\vh}_\ell^t(\vx)
    +
    \left(
        \bar{\vh}_\ell^t(\vx) \mU_{\mathrm{in},\ell+1}^t
    \right)
    \mC_\ell^t
    \left(\mV_{\mathrm{out},\ell}^t\right)^\top.
    \end{aligned}
    \label{eq:bridge_correction}
\end{equation}
The corrected representation $\bar{\vh}_\ell^{\prime,t}(\vx)$ is then passed to merged block $\ell+1$, so corrections are propagated sequentially through the backbone.
At the terminal location, the analogous update using $\mB_L^t$ produces the calibrated final representation $\bar{\vh}_L^{\prime,t}(\vx)$.

Let $\vh_\ell^t(\vx)$ denote the corresponding expert representation.
We jointly optimize all \Bridge{} cores for task $t$ by minimizing the feature discrepancy across calibration locations:
\begin{equation}
    \min_{\{\mC_\ell^t\}_{\ell=1}^{L}}
    \sum_{\ell=1}^{L}
    \mathbb E_{\vx\sim\mathcal D_t}
    \left[
        \left\|\bar{\vh}_\ell^{\prime,t}(\vx)-\vh_\ell^t(\vx)\right\|_F^2
    \right].
    \label{eq:calibration_loss}
\end{equation}
Since the frozen expert provides the target representations, calibration requires no task labels.
\section{Experimental Results}
\label{sec:experiments}
We compare \ESCal{} with existing post-merging calibration methods
under different base merging methods.
Our evaluation examines both the improvement over uncalibrated
merged models and the remaining performance gap to the expert models.
We conduct our main evaluation on vision and natural language
understanding (NLU), and additionally evaluate natural language
generation (NLG) to examine the applicability of \ESCal{} beyond
these settings. 
The NLG results and discussion are provided in Appendix~\ref{app:nlg_results}.

\paragraph{Tasks and Models.}
For vision, we use CLIP ViT-B/32, ViT-B/16, and ViT-L/14
models~\citep{radford2021clipvit} fine-tuned on eight image
classification tasks.
For NLU, we use RoBERTa-base~\citep{liu2019roberta} models fine-tuned on eight GLUE tasks~\citep{wang2018glue}.
In both settings, we merge the eight task-specific models and evaluate the resulting model across all tasks.
For NLG, we use three LLaMA-2-13B~\citep{touvron2023llama} experts
specialized for instruction following, mathematical reasoning, and code
generation, and consider all pairwise merges.

\paragraph{Merging and Calibration Methods.}
We consider \TA{}~\citep{ilharco2023ta},
\TIES{}~\citep{yadav2023ties},
\AdaMerging{}~\citep{yang2024adamerging},
\WUDI{}~\citep{cheng2025wudi}, and
\ProDistill{}~\citep{xu2025prodistill}
as base merging methods.
For post-merging calibration, we compare
\Surgery{}~\citep{yang2024surgery},
\Surgery{V2}~\citep{yang2024surgeryv2},
\IntervMerge{}~\citep{osial2024interv},
\FeatCal{}~\citep{gu2026featcal}, and \ESCal{} (ours).
We report results for each base merged model before and after calibration.
We also include the pretrained model and task-specific experts as references.

\paragraph{Calibration Protocol.}
For each base merging method, all calibration methods start from the same merged checkpoint and use the same task-specific calibration sets, each containing 64 input samples.
\Surgery{}, \Surgery{V2}, \IntervMerge{}, and \ESCal{} optimize only their calibration modules with the merged weights frozen, whereas \FeatCal{} applies closed-form updates to the merged weights.
Calibration and evaluation sets are disjoint, and results are averaged over five random seeds unless otherwise specified.
For \IntervMerge{}, we select settings separately for each architecture and base using the existing seed-0 search scores (vision test accuracy and NLU GLUE validation score), then fix these settings across seeds 0--4.
Detailed hyperparameter settings are provided in Appendix~\ref{app:hyperparameter_search}.

\subsection{Results on Vision Tasks}
Table~\ref{tab:taskwise_accuracy_b32} presents representative ViT-B/32
results with \TA{}, \TIES{}, and \WUDI{}.
Complete results across architectures and base merging methods are
provided in Appendix~\ref{app:vision_results}.

\begin{table*}[htbp]
\centering
\renewcommand{\arraystretch}{0.94}
\caption{\textbf{Vision model merging results on ViT-B/32.}
Task-wise test accuracy (\%) and average performance across eight image
classification tasks.}
\label{tab:taskwise_accuracy_b32}
\renewcommand{\arraystretch}{1.1}
\resizebox{0.9\textwidth}{!}{
\begin{tabular}{lcccccccc|c}
\toprule
\textbf{Method}
& \textbf{SUN397} & \textbf{Cars} & \textbf{RESISC45} & \textbf{EuroSAT}
& \textbf{SVHN} & \textbf{GTSRB} & \textbf{MNIST} & \textbf{DTD} & \textbf{Avg.} \\
\midrule
Pretrained
& 63.18 & 59.73 & 60.33 & 45.37 & 31.61
& 32.56 & 48.25 & 43.99 & 48.13 \\
Expert
& 75.34 & 77.73 & 95.98 & 99.74 & 97.46
& 98.73 & 99.69 & 79.36 & 90.50 \\
\midrule

\TA{}
& 55.32 & 54.98 & 66.82 & 77.78 & 80.21 & 69.68 & 97.34 & 50.37 & 69.06 \\
\quad w/ \Surgery{}
& 60.39 & 52.62 & 71.38 & 88.65 & 81.67 & 74.27 & 97.82 & 52.77 & 72.45 \\
\quad w/ \Surgery{V2}
& \textbf{70.70} & 66.41 & 90.65 & 99.12 & 96.61 & 95.88 & 99.47 & 70.90 & 86.22 \\
\quad w/ \IntervMerge{}
& 66.35 & 58.87 & 85.88 & 97.61 & 93.44 & 87.49 & 98.62 & 64.31 & 81.57 \\
\quad w/ \FeatCal{}
& 69.22 & \textbf{70.31} & 89.06 & 98.00 & 95.01 & 92.40 & 99.13 & 71.11 & 85.53 \\
\quad \textbf{w/ \ESCal}
& 70.33 & 69.02 & \textbf{93.60} & \textbf{99.48} & \textbf{97.10} & \textbf{97.70} & \textbf{99.55} & \textbf{73.52} & \textbf{87.54} \\

\midrule

\TIES{}
& 65.05 & 64.43 & 74.92 & 76.48 & 81.28 & 69.37 & 96.53 & 54.52 & 72.82 \\
\quad w/ \Surgery{}
& 67.64 & 59.57 & 78.50 & 89.24 & 81.30 & 76.07 & 97.14 & 60.03 & 76.19 \\
\quad w/ \Surgery{V2}
& 70.87 & 66.26 & 90.14 & 98.94 & 96.54 & 95.52 & 99.45 & 70.89 & 86.08 \\
\quad w/ \IntervMerge{}
& 68.65 & 60.59 & 86.21 & 97.39 & 92.99 & 87.60 & 98.24 & 65.14 & 82.10 \\
\quad w/ \FeatCal{}
& 70.12 & \textbf{70.74} & 89.28 & 97.93 & 94.71 & 92.55 & 99.10 & 71.13 & 85.69 \\
\quad \textbf{w/ \ESCal}
& \textbf{70.93} & 69.23 & \textbf{93.31} & \textbf{99.36} & \textbf{97.09} & \textbf{97.70} & \textbf{99.53} & \textbf{73.35} & \textbf{87.56} \\

\midrule

\WUDI{}
& 69.11 & 71.38 & 85.62 & 95.63 & 94.27 & 94.51 & 99.47 & 69.47 & 84.93 \\
\quad w/ \Surgery{}
& 69.24 & 69.42 & 87.70 & 97.53 & 94.81 & 95.43 & 99.48 & 71.37 & 85.62 \\
\quad w/ \Surgery{V2}
& 73.24 & 73.57 & 94.15 & \textbf{99.59} & 97.17 & 98.38 & \textbf{99.64} & \textbf{77.66} & 89.18 \\
\quad w/ \IntervMerge{}
& 71.72 & 72.55 & 92.65 & 99.21 & 96.33 & 97.97 & 99.59 & 75.89 & 88.24 \\
\quad w/ \FeatCal{}
& 71.98 & \textbf{75.80} & 93.71 & 99.31 & 96.53 & 98.42 & 99.50 & 77.52 & 89.10 \\
\quad \textbf{w/ \ESCal}
& \textbf{73.55} & 74.68 & \textbf{94.80} & 99.58 & \textbf{97.31} & \textbf{98.68} & \textbf{99.64} & 77.52 & \textbf{89.47} \\

\bottomrule
\end{tabular}
}
\end{table*}

\ESCal{} improves the average accuracy of all three base merged models
shown in the table and achieves the highest average among the compared
calibration methods.
The largest improvement over the uncalibrated model occurs with \TA{},
whose average accuracy rises from $69.06\%$ to $87.54\%$, a gain of
$18.48$ percentage points.
\ESCal{} also improves \WUDI{}, which has the highest uncalibrated
average accuracy among the three base merging methods, from $84.93\%$
to $89.47\%$, a gain of $4.54$ percentage points.
These results show that calibration remains beneficial even when the
base merged model already achieves high accuracy.

These gains extend to both task-wise and layer-wise \AdaMerging{}
(Appendix~\ref{app:vision_results}).
On ViT-B/32, \ESCal{} increases their average accuracies from
$70.82\%$ to $86.87\%$ and from $79.31\%$ to $87.91\%$, respectively,
showing that calibration can provide further gains after merging
coefficients have been adapted.
The improvements also hold for \ProDistill{} across all three architectures.
Thus, \ESCal{} can complement merging methods that already use
intermediate-feature distillation.

\subsection{Results on NLU Tasks}

Table~\ref{tab:roberta_nlu} reports representative results with
\TA{}, \TIES{}, and \WUDI{}.
Complete results across base merging methods are provided in
Appendix~\ref{app:nlu_results}.

\begin{table}[htbp]
    \centering
\renewcommand{\arraystretch}{0.94}
    \caption{\textbf{NLU model merging results on RoBERTa-base.}
Task-wise performance and average score across eight GLUE tasks.}
    \label{tab:roberta_nlu}
    \renewcommand{\arraystretch}{1.1}
    \resizebox{0.9\linewidth}{!}{
    \begin{tabular}{lcccccccc|c}
        \toprule
        \textbf{Method}
        & \textbf{CoLA}
        & \textbf{SST-2}
        & \textbf{MRPC}
        & \textbf{STS-B}
        & \textbf{QQP}
        & \textbf{MNLI}
        & \textbf{QNLI}
        & \textbf{RTE}
        & \textbf{Avg.} \\
        \midrule
        Pretrained
        & 0.0000 & 0.5975 & 0.1581 & -0.0758 & 0.4633
        & 0.3183 & 0.5078 & 0.5271 & 0.3120 \\

        Expert
        & 0.5895 & 0.9438 & 0.9091 & 0.8997 & 0.8994
        & 0.8718 & 0.9198 & 0.7726 & 0.8507 \\
        \midrule

        \TA{}
        & 0.1232 & 0.6135 & 0.5500 & 0.5356 & 0.5881 & 0.6270 & 0.5365 & 0.5343 & 0.5135 \\
        \quad w/ \Surgery{}
        & 0.2553 & 0.8532 & 0.7657 & 0.7877 & 0.7793 & 0.7215 & 0.7839 & 0.7184 & 0.7081 \\
        \quad w/ \Surgery{V2}
        & 0.5358 & \textbf{0.9326} & 0.8908 & 0.8978 & 0.8002 & 0.8404 & 0.9084 & 0.7408 & 0.8184 \\
        \quad w/ \IntervMerge{}
& 0.4647 & 0.9135 & 0.8477 & 0.8706 & 0.7779 & 0.7625 & 0.8524 & 0.7199 & 0.7762 \\
        \quad w/ \FeatCal{}
        & 0.3879 & 0.9140 & 0.8162 & 0.4900 & 0.7680 & 0.7569 & 0.8499 & 0.7011 & 0.7105 \\
        \quad \textbf{w/ \ESCal}
        & \textbf{0.5588} & 0.9319 & \textbf{0.8937} & \textbf{0.8998} & \textbf{0.8061} & \textbf{0.8469} & \textbf{0.9117} & \textbf{0.7415} & \textbf{0.8238} \\

        \midrule

        \TIES{}
        & 0.1972 & 0.7626 & 0.5075 & 0.4390 & 0.6039 & 0.6670 & 0.5380 & 0.4946 & 0.5262 \\
        \quad w/ \Surgery{}
        & 0.2680 & 0.8635 & 0.7570 & 0.7770 & 0.7770 & 0.7347 & 0.7988 & 0.6989 & 0.7094 \\
        \quad w/ \Surgery{V2}
        & 0.5420 & 0.9319 & \textbf{0.8848} & \textbf{0.8965} & \textbf{0.8205} & \textbf{0.8474} & \textbf{0.9079} & 0.7394 & \textbf{0.8213} \\
        \quad w/ \IntervMerge{}
& 0.4382 & 0.9174 & 0.8368 & 0.8702 & 0.8023 & 0.7805 & 0.8660 & 0.7249 & 0.7796 \\
        \quad w/ \FeatCal{}
        & 0.4055 & 0.9106 & 0.8111 & 0.4765 & 0.7702 & 0.7569 & 0.8433 & 0.7061 & 0.7100 \\
        \quad \textbf{w/ \ESCal}
        & \textbf{0.5552} & \textbf{0.9339} & 0.8769 & 0.8935 & 0.8144 & 0.8443 & 0.9070 & \textbf{0.7430} & 0.8210 \\

        \midrule

        \WUDI{}
        & 0.0625 & 0.8073 & 0.2565 & 0.0927 & 0.7680 & 0.4431 & 0.4622 & 0.4874 & 0.4225 \\
        \quad w/ \Surgery{}
        & 0.2226 & 0.9025 & 0.7860 & 0.7758 & 0.8270 & 0.7755 & 0.8118 & 0.7206 & 0.7277 \\
        \quad w/ \Surgery{V2}
        & 0.5529 & 0.9392 & 0.8908 & 0.8868 & \textbf{0.8798} & \textbf{0.8620} & 0.9136 & \textbf{0.7603} & 0.8357 \\
        \quad w/ \IntervMerge{}
& 0.4766 & 0.9282 & 0.8303 & 0.8668 & 0.8562 & 0.8347 & 0.8884 & 0.7516 & 0.8041 \\
        \quad w/ \FeatCal{}
        & 0.4387 & 0.9190 & 0.8215 & 0.4953 & 0.8234 & 0.7748 & 0.8358 & 0.6888 & 0.7247 \\
        \quad \textbf{w/ \ESCal}
        & \textbf{0.5663} & \textbf{0.9397} & \textbf{0.8950} & \textbf{0.8953} & 0.8744 & 0.8616 & \textbf{0.9144} & 0.7545 & \textbf{0.8376} \\

        \bottomrule
    \end{tabular}
    }
\end{table}

\ESCal{} improves the average GLUE score for all three base merging methods.
\WUDI{} benefits most, with its average score rising from $0.4225$
to $0.8376$.
Despite having the lowest uncalibrated score among the three methods,
\WUDI{} achieves the highest average after \ESCal{} calibration.
Across these settings, \ESCal{} consistently outperforms \Surgery{} and \FeatCal{}.
Relative to \Surgery{V2}, it achieves higher averages with \TA{}
and \WUDI{}, and a comparable average with \TIES{}.

The improvements also extend to the additional base merging methods in
Appendix~\ref{app:nlu_results}.
For task-wise and layer-wise \AdaMerging{}, \ESCal{} raises the average
scores from $0.4989$ to $0.8202$ and from $0.4823$ to $0.8276$,
respectively; for \ProDistill{}, it improves the average from $0.7298$
to $0.7999$.
These results support the effectiveness of error-subspace calibration
across merged checkpoints produced by different merging strategies.

\section{Further Analyses}
We further analyze \ESCal{} to examine how its error-subspace
design contributes to calibration performance and how closely
the calibrated model matches expert predictions.
We also assess its calibration data and parameter requirements,
along with the additional storage and inference overhead
of its calibration modules.

\subsection{Effectiveness of Error-Subspace Construction}
\label{sec:error_subspace_construction}
\begin{table}[!htbp]
\centering
\caption{
We compare different basis sources on ViT-B/32 merged with \TA{}
using rank $r=128$.
We report mean accuracy here, with standard deviations provided in Table~\ref{tab:vision_ablation_basis_full} in Appendix~\ref{app:error_subspace_details}.}
\label{tab:vision_ablation_basis}
\renewcommand{\arraystretch}{1.15}
\resizebox{\textwidth}{!}{
\begin{tabular}{lcccccccc|c}
\toprule
\textbf{Basis}
& \textbf{SUN397}
& \textbf{Cars}
& \textbf{RESISC45}
& \textbf{EuroSAT}
& \textbf{SVHN}
& \textbf{GTSRB}
& \textbf{MNIST}
& \textbf{DTD}
& \textbf{Avg.} \\
\midrule
Random orthonormal
& 65.42 & 63.88 & 88.54
& 98.61 & 96.06 & 95.81
& 99.32 & 64.98 & 84.08 \\
Fine-tuned weights, top-$r$
& 66.94 & 65.02 & 89.85
& 99.03 & 96.17 & 96.32
& 99.36 & 65.70 & 84.80 \\
Task vector, top-$r$
& 69.29 & 67.72 & 93.01
& 99.44 & 97.08 & 97.47
& 99.53 & 72.53 & 87.01 \\
\textbf{Merging error, top-$r$}
& \textbf{70.33} & \textbf{69.02} & \textbf{93.60}
& \textbf{99.48} & \textbf{97.10} & \textbf{97.70}
& \textbf{99.55} & \textbf{73.52} & \textbf{87.54} \\
\bottomrule
\end{tabular}
}
\end{table}

\paragraph{Source of calibration subspaces.}

We first compare top-$r$ singular subspaces derived from fine-tuned
weights, task vectors, and merging errors, together with a random
orthonormal baseline.
We construct the random baseline by sampling Gaussian random matrices
and applying reduced QR decomposition.

As shown in Table~\ref{tab:vision_ablation_basis}, error-derived bases
achieve the highest average accuracy of $87.54\%$, followed by
task-vector bases with $87.01\%$ and fine-tuned-weight bases with
$84.80\%$.
Fine-tuned weights represent the full expert transformation and do not
explicitly isolate the parameter changes introduced by fine-tuning.
Task vectors capture these changes relative to the pretrained model,
whereas merging errors isolate the discrepancy between the
expert and the merged model being calibrated.
The observed ordering, therefore, supports constructing calibration
subspaces from the discrepancy that the calibration is intended to
correct.

\begin{table}[!htbp]
\centering
\caption{
We independently vary the read and write directions for \ESCal{}
with \TA{} on ViT-B/32 using rank $r=128$.
Random-$r$ denotes $r$ directions selected from the singular-vector basis of the merging error.
}

\label{tab:abl_read_write_direction}

\small
\renewcommand{\arraystretch}{1.1}
\resizebox{0.55\textwidth}{!}{
\begin{tabular}{lccc}
\toprule
\multirow{2}{*}{\textbf{Read basis}}
& \multicolumn{3}{c}{\textbf{Write basis}} \\
\cmidrule(lr){2-4}
& Top-$r$
& Random-$r$
& Bottom-$r$ \\
\midrule
Top-$r$
& $\mathbf{87.54 \pm 0.07}$
& $84.24 \pm 0.18$
& $81.69 \pm 0.12$ \\
Random-$r$
& $85.96 \pm 0.15$
& $82.93 \pm 0.22$
& $80.89 \pm 0.19$ \\
Bottom-$r$
& $85.35 \pm 0.07$
& $82.63 \pm 0.20$
& $80.86 \pm 0.13$ \\
\bottomrule
\end{tabular}
}
\end{table}

\paragraph{Selection of read and write directions.}
We next examine which directions within the merging-error subspaces are most effective for reading and writing corrections.
Table~\ref{tab:abl_read_write_direction} independently varies the read and write bases among the top-$r$, random-$r$, and bottom-$r$ singular directions of the merging error.
Unlike the random orthonormal baseline considered above, Random-$r$ selects $r$ directions from the singular vectors of the merging error.
Using the top-$r$ singular directions for both the read and write bases achieves the highest average accuracy of $87.54\%$.

We find that the choice of write basis has a larger effect on performance.
For a fixed read basis, replacing the top-$r$ write directions with random-$r$ or bottom-$r$ directions consistently causes substantial degradation.
This suggests that corrections should be written along the dominant output-error directions rather than random or bottom singular directions.

We also find that, for a fixed write basis, the top-$r$ read directions consistently perform best, while the bottom-$r$ directions perform worst.
This indicates that, beyond correcting directions in which the current representation deviates from the expert, reading from input directions that most strongly affect the following transformation's local output discrepancy is beneficial.
Together, these results support the \ESCal{} design of writing along the dominant current output-error directions while reading from the dominant input-error directions of the next transformation.

\begin{table}[!htbp]
\centering
\caption{Constrained uses a fixed top-$r$ error basis. Unconstrained learns the corresponding projection from a random initialization without restricting its directions to the error subspace.}
\renewcommand{\arraystretch}{1.15}
\label{tab:abl_subspace_constraint}
\small
\begin{tabular}{llccc}
\toprule
\textbf{Read} &
\textbf{Write} &
\textbf{$r=64$} &
\textbf{$r=128$} &
\textbf{$r=256$} \\
\midrule
Constrained & Constrained
& $86.91 \pm 0.06$ & $87.54 \pm 0.07$ & $87.62 \pm 0.12$ \\

Unconstrained & Constrained
& $87.23 \pm 0.05$ & $87.50 \pm 0.04$ & $87.62 \pm 0.14$ \\

Constrained & Unconstrained
& $87.53 \pm 0.10$ & $87.76 \pm 0.04$ & $87.75 \pm 0.07$ \\
\bottomrule
\end{tabular}
\end{table}

\paragraph{Role of the read- and write-subspace constraints.}
The preceding ablations examine which basis sources and singular directions are most effective when the read and write bases are fixed.
We next examine whether relaxing either subspace constraint provides
additional gains by replacing the corresponding fixed basis with a
learnable projection.
Specifically, we make either the read or write projection learnable
from a random initialization while keeping the other constrained to
its fixed top-$r$ error basis.

As shown in Table~\ref{tab:abl_subspace_constraint}, relaxing the write constraint improves accuracy across all three ranks, whereas relaxing the read constraint provides a gain only at $r=64$.
The more consistent gains from adapting the write projection further highlight the importance of the directions along which corrections are applied.
However, relaxing either constraint requires learning a $d\times r$ projection in place of the corresponding fixed basis, increasing the number of trainable parameters for that projection by a factor of $d/r$ compared with the $r\times r$ core.
Given the modest performance gains relative to this additional parameter cost, we keep both bases fixed in \ESCal{}.

\subsection{Prediction Alignment with Expert Models}
\label{sec:prediction_alignment}
To complement task-performance evaluation, we measure how closely the
calibrated merged models match expert predictions using KL divergence
and prediction agreement.
Following the standard evaluation protocol for multi-task model merging,
we pair each calibrated merged backbone with the corresponding fixed expert
head and compare its predictions with those of the expert.
For each task, we compute KL divergence from the expert predictive distribution to that of the calibrated merged model and prediction agreement, defined as the fraction of matching predicted labels, and report the arithmetic mean across tasks for KL divergence and the sample-count-weighted mean across tasks for prediction agreement. For RoBERTa, we compute these metrics over the seven classification tasks, excluding STS-B, which is a regression task.

\begin{table*}[htbp]
\centering
\caption{\textbf{Prediction alignment with expert models.}
KL divergence and prediction agreement are reported across vision and NLU architectures.}
\label{tab:prediction_alignment}
\renewcommand{\arraystretch}{1.1}
\resizebox{\textwidth}{!}{
\begin{tabular}{lcccccccc}
\toprule
\multirow{2}{*}{Method}
& \multicolumn{2}{c}{ViT-B/32}
& \multicolumn{2}{c}{ViT-B/16}
& \multicolumn{2}{c}{ViT-L/14}
& \multicolumn{2}{c}{RoBERTa} \\
\cmidrule(lr){2-3}
\cmidrule(lr){4-5}
\cmidrule(lr){6-7}
\cmidrule(lr){8-9}
& KL $\downarrow$ & Agreement $\uparrow$
& KL $\downarrow$ & Agreement $\uparrow$
& KL $\downarrow$ & Agreement $\uparrow$
& KL $\downarrow$ & Agreement $\uparrow$ \\
\midrule
\TA{}
& 0.9383 & 74.51\%
& 0.7918 & 79.56\%
& 0.4447 & 86.83\%
& 0.6926 & 61.38\% \\

\quad w/ \Surgery{V2}
& 0.2671 & 90.70\%
& 0.2054 & 92.94\%
& 0.1259 & 95.27\%
& 0.2872 & 87.76\% \\

\quad w/ \FeatCal{}
& 0.3204 & 89.48\%
& 0.2987 & 90.96\%
& 0.1610 & 94.40\%
& 0.4697 & 83.35\% \\

\quad w/ \textbf{\ESCal}
& \textbf{0.1821} & \textbf{92.09\%}
& \textbf{0.1341} & \textbf{94.16\%}
& \textbf{0.0910} & \textbf{96.09\%}
& \textbf{0.2439} & \textbf{88.74\%} \\
\bottomrule
\end{tabular}
}
\end{table*}

With \TA{} as the base merging method, \ESCal{} achieves the lowest KL
divergence and the highest prediction agreement among the compared methods
across all four evaluated architectures.
On ViT-B/32, for example, relative to \Surgery{V2}, \ESCal{} reduces
KL divergence from $0.2671$ to $0.1821$ and increases prediction agreement
from $90.70\%$ to $92.09\%$.
These results show that \ESCal{} more closely matches expert predictions,
with stronger alignment in both predictive distributions and predicted labels.

\subsection{Calibration Efficiency and Overhead}
\label{sec:train_efficiency}

We examine the parameter efficiency and calibration data requirements of \ESCal{} on ViT-B/32 with \TA{}.
We also assess the additional storage and inference overhead associated with its calibration modules.

\begin{table}[htbp]
\centering
\captionsetup{justification=centering}
\setlength{\tabcolsep}{5pt}
\begin{minipage}[t]{0.44\linewidth}
    \centering
    \caption{Calibration cost and accuracy.}
    \label{tab:calibration_cost}
    \small
    \begin{tabular}{lccc}
    \toprule
    Method
    & \shortstack{Trainable\\params.}
    & \shortstack{Runtime\\(min)}
    & \shortstack{Avg. acc.\\(\%)} \\
    \midrule
    \Surgery{}
    & 131,072 & 8.48 & 72.45 \\
    \Surgery{V2}
    & 2,490,368 & 13.76 & 86.22 \\
    \FeatCal{}
    & 0 & 0.63 & 85.53 \\
    \textbf{\ESCal{}}
    & 1,572,864 & 13.22 & \textbf{87.54} \\

    \bottomrule
    \end{tabular}
\end{minipage}\hspace{0.025\linewidth}%
\begin{minipage}[t]{0.27\linewidth}
    \centering
    \caption{Calibration budget.}
    \label{tab:calibration_budget}
    \small
    \setlength{\tabcolsep}{8pt}
    \begin{tabular}{cc}
    \toprule
    \shortstack{Examples\\per task}
    & \shortstack{Avg. acc.\\(\%)} \\
    \midrule
    32 & $86.48 \pm 0.21$ \\
    64 & $87.54 \pm 0.07$ \\
    128 & $88.07 \pm 0.06$ \\
    256 & $88.29 \pm 0.07$ \\
    \bottomrule
    \end{tabular}
\end{minipage}
\end{table}

\paragraph{Parameter efficiency and inference overhead.}
\ESCal{} achieves higher average accuracy with $36.8\%$ fewer trainable parameters than \Surgery{V2} on ViT-B/32 (Table~\ref{tab:calibration_cost}), while requiring comparable calibration runtime under the selected configurations.
For \FeatCal{}, we report zero trainable parameters because its calibration updates are computed in closed form without gradient-based optimization.
While only the core mappings are learned, the fixed read/write bases must also be retained at inference time, requiring additional storage beyond the trainable parameters.
For \TA{} on ViT-B/32 with $r=128$, the bases and cores together add $2.91\%$ per task to the shared image encoder's parameter storage at the same precision.
We additionally measure the inference overhead relative to the uncalibrated model in terms of latency and throughput, and report these results together with detailed storage accounting in Appendix~\ref{app:computational_cost}.

\paragraph{Calibration data budget.}
We vary the number of calibration examples per task from 4 to 256 for
\ESCal{} with \TA{} on ViT-B/32 and evaluate on the full test sets.
Table~\ref{tab:calibration_budget} reports the results from 32 to 256
examples, with the full sweep provided in Appendix~\ref{app:sample_efficiency}.
Our default budget of 64 examples per task provides a favorable trade-off
between accuracy and calibration data requirements.
\section{Conclusion and Limitations}
We introduced \ESCal{}, a post-merging calibration framework that restricts feature correction to low-dimensional error subspaces derived from weight differences between expert and merged models.
Its \Bridge{} modules learn only small linear mappings between fixed subspaces, enabling parameter-efficient calibration.
Across vision and natural language understanding, our results show that these error subspaces provide an effective structure for recovering task-specific capabilities after merging.
This highlights the importance of identifying \emph{which directions} to correct, in addition to optimizing representation alignment.

\paragraph{Limitations \& Future Directions.}
First, each \Bridge{} learns a low-rank linear correction to the blockwise hidden-representation gap between the merged model and an expert, with both subspace bases fixed.
This design draws on prior work on lightweight post-merging calibration~\citep{yang2024surgery,yang2024surgeryv2}.
Our experiments demonstrate its effectiveness in the evaluated settings, but its general applicability remains unverified.
Thus, \ESCal{} may be less effective when the required corrections fall outside the fixed error subspaces or require nonlinear transformations.

Second, the error-subspace dimension $r$ must be chosen before \Bridge{} training and may need to vary across architectures and domains.
We treat $r$ as a hyperparameter, but do not systematically investigate its relationship with the correction capacity of \Bridge{} modules.
We leave this analysis and adaptive rank allocation to future work.

Third, \ESCal{} retains its fixed read/write bases at inference time, requiring additional storage beyond the learned \Bridge{} core modules.
However, this overhead is relatively small.
For example, with \TA{} on ViT-B/32 and $r=128$, the bases and cores together add $2.91\%$ per task to the shared image encoder's parameter storage at the same precision (Table~\ref{tab:parameter_storage_cost}).

Finally, we study \Bridge{} modules only at block boundaries.
Exploring alternative placements, including within blocks, could help clarify how correction placement affects representation alignment.
More expressive correction modules also remain an architectural direction for future work.

\clearpage

\bibliography{paper}

\begin{thebibliography}{29}
\providecommand{\natexlab}[1]{#1}
\providecommand{\url}[1]{\texttt{#1}}
\expandafter\ifx\csname urlstyle\endcsname\relax
  \providecommand{\doi}[1]{doi: #1}\else
  \providecommand{\doi}{doi: \begingroup \urlstyle{rm}\Url}\fi

\bibitem[Cheng et~al.(2017)Cheng, Han, and Lu]{cheng2017resics}
Gong Cheng, Junwei Han, and Xiaoqiang Lu.
\newblock Remote sensing image scene classification: Benchmark and state of the art.
\newblock \emph{Proceedings of the IEEE}, 105\penalty0 (10):\penalty0 1865--1883, 2017.

\bibitem[Cheng et~al.(2025)Cheng, Xiong, Wei, Zhu, and Yuan]{cheng2025wudi}
Runxi Cheng, Feng Xiong, Yongxian Wei, Wanyun Zhu, and Chun Yuan.
\newblock Whoever started the interference should end it: Guiding data-free model merging via task vectors.
\newblock In \emph{Forty-second International Conference on Machine Learning}, 2025.
\newblock URL \url{https://openreview.net/forum?id=xR9msNaREW}.

\bibitem[Cimpoi et~al.(2014)Cimpoi, Maji, Kokkinos, Mohamed, and Vedaldi]{cimpoi2014dtd}
Mircea Cimpoi, Subhransu Maji, Iasonas Kokkinos, Sammy Mohamed, and Andrea Vedaldi.
\newblock Describing textures in the wild.
\newblock In \emph{Proceedings of the IEEE conference on computer vision and pattern recognition}, pages 3606--3613, 2014.

\bibitem[Gargiulo et~al.(2025)Gargiulo, Crisostomi, Bucarelli, Scardapane, Silvestri, and Rodola]{gargiulo2025tsv}
Antonio~Andrea Gargiulo, Donato Crisostomi, Maria~Sofia Bucarelli, Simone Scardapane, Fabrizio Silvestri, and Emanuele Rodola.
\newblock Task singular vectors: Reducing task interference in model merging.
\newblock In \emph{2025 IEEE/CVF Conference on Computer Vision and Pattern Recognition (CVPR)}, pages 18695--18705. IEEE, 2025.

\bibitem[Gu et~al.(2026)Gu, Cai, Wang, Wang, Wang, Wang, Huang, Lu, Wu, and Yang]{gu2026featcal}
Yanggan Gu, Shuo Cai, Zihao Wang, Wenjun Wang, Yuanyi Wang, Pengkai Wang, Sirui Huang, Su~Lu, Jianmin Wu, and Hongxia Yang.
\newblock {FeatCal}: Feature calibration for post-merging models.
\newblock \emph{arXiv preprint arXiv:2605.13030}, 2026.

\bibitem[Helber et~al.(2019)Helber, Bischke, Dengel, and Borth]{helber2019eurosat}
Patrick Helber, Benjamin Bischke, Andreas Dengel, and Damian Borth.
\newblock {EuroSAT}: A novel dataset and deep learning benchmark for land use and land cover classification.
\newblock \emph{IEEE Journal of Selected Topics in Applied Earth Observations and Remote Sensing}, 12\penalty0 (7):\penalty0 2217--2226, 2019.

\bibitem[Ilharco et~al.(2023)Ilharco, Ribeiro, Wortsman, Schmidt, Hajishirzi, and Farhadi]{ilharco2023ta}
Gabriel Ilharco, Marco~Tulio Ribeiro, Mitchell Wortsman, Ludwig Schmidt, Hannaneh Hajishirzi, and Ali Farhadi.
\newblock Editing models with task arithmetic.
\newblock In \emph{The Eleventh International Conference on Learning Representations}, 2023.
\newblock URL \url{https://openreview.net/forum?id=6t0Kwf8-jrj}.

\bibitem[Krause et~al.(2013)Krause, Stark, Deng, and Fei-Fei]{krause2013cars}
Jonathan Krause, Michael Stark, Jia Deng, and Li~Fei-Fei.
\newblock {3D} object representations for fine-grained categorization.
\newblock In \emph{Proceedings of the IEEE international conference on computer vision workshops}, pages 554--561, 2013.

\bibitem[LeCun et~al.(1998)LeCun, Bottou, Bengio, and Haffner]{lecun1998mnist}
Yann LeCun, L{\'e}on Bottou, Yoshua Bengio, and Patrick Haffner.
\newblock Gradient-based learning applied to document recognition.
\newblock \emph{Proceedings of the IEEE}, 86\penalty0 (11):\penalty0 2278--2324, 1998.

\bibitem[Liu et~al.(2019)Liu, Ott, Goyal, Du, Joshi, Chen, Levy, Lewis, Zettlemoyer, and Stoyanov]{liu2019roberta}
Yinhan Liu, Myle Ott, Naman Goyal, Jingfei Du, Mandar Joshi, Danqi Chen, Omer Levy, Mike Lewis, Luke Zettlemoyer, and Veselin Stoyanov.
\newblock {RoBERTa}: A robustly optimized {BERT} pretraining approach.
\newblock \emph{arXiv preprint arXiv:1907.11692}, 2019.

\bibitem[Luo et~al.(2025)Luo, Sun, Xu, Zhao, Lou, Tao, Geng, Lin, Chen, Tang, et~al.]{luo2025wizardmath}
Haipeng Luo, Qingfeng Sun, Can Xu, Pu~Zhao, Jian-Guang Lou, Chongyang Tao, Xiubo Geng, Qingwei Lin, Shifeng Chen, Yansong Tang, et~al.
\newblock {WizardMath}: Empowering mathematical reasoning for large language models via reinforced evol-instruct.
\newblock In \emph{International Conference on Learning Representations}, pages 49573--49609, 2025.

\bibitem[Marczak et~al.(2025)Marczak, Magistri, Cygert, Twardowski, Bagdanov, and van~de Weijer]{marczak2025isomerging}
Daniel Marczak, Simone Magistri, Sebastian Cygert, Bart{\l}omiej Twardowski, Andrew~D. Bagdanov, and Joost van~de Weijer.
\newblock No task left behind: Isotropic model merging with common and task-specific subspaces.
\newblock In \emph{Forty-second International Conference on Machine Learning}, 2025.
\newblock URL \url{https://openreview.net/forum?id=RBZpAa27ls}.

\bibitem[Netzer et~al.(2011)Netzer, Wang, Coates, Bissacco, Wu, and Ng]{netzer2011svhn}
Yuval Netzer, Tao Wang, Adam Coates, Alessandro Bissacco, Baolin Wu, and Andrew~Y Ng.
\newblock Reading digits in natural images with unsupervised feature learning.
\newblock In \emph{NIPS workshop on deep learning and unsupervised feature learning}, number~2, page~4. Granada, 2011.

\bibitem[Osial et~al.(2024)Osial, Marczak, et~al.]{osial2024interv}
Marcin Osial, Daniel Marczak, et~al.
\newblock Parameter-efficient interventions for enhanced model merging.
\newblock \emph{arXiv preprint arXiv:2412.17023}, 2024.

\bibitem[Radford et~al.(2021)Radford, Kim, Hallacy, Ramesh, Goh, Agarwal, Sastry, Askell, Mishkin, Clark, et~al.]{radford2021clipvit}
Alec Radford, Jong~Wook Kim, Chris Hallacy, Aditya Ramesh, Gabriel Goh, Sandhini Agarwal, Girish Sastry, Amanda Askell, Pamela Mishkin, Jack Clark, et~al.
\newblock Learning transferable visual models from natural language supervision.
\newblock In \emph{International conference on machine learning}, pages 8748--8763. PMLR, 2021.

\bibitem[Skorobogat et~al.(2025)Skorobogat, Roth, and Georgescu]{skorobogat2025subspace}
Ronald Skorobogat, Karsten Roth, and Mariana-Iuliana Georgescu.
\newblock Subspace-boosted model merging.
\newblock \emph{arXiv preprint arXiv:2506.16506}, 2025.

\bibitem[Stallkamp et~al.(2011)Stallkamp, Schlipsing, Salmen, and Igel]{stallkamp2011gtsrb}
Johannes Stallkamp, Marc Schlipsing, Jan Salmen, and Christian Igel.
\newblock The {German} traffic sign recognition benchmark: a multi-class classification competition.
\newblock In \emph{The 2011 international joint conference on neural networks}, pages 1453--1460. IEEE, 2011.

\bibitem[Stoica et~al.(2025)Stoica, Ramesh, Ecsedi, Choshen, and Hoffman]{stoica2025knots}
George Stoica, Pratik Ramesh, Boglarka Ecsedi, Leshem Choshen, and Judy Hoffman.
\newblock Model merging with {SVD} to tie the knots.
\newblock In \emph{International Conference on Learning Representations}, volume 2025, pages 4501--4519, 2025.

\bibitem[Touvron et~al.(2023)Touvron, Martin, Stone, Albert, Almahairi, Babaei, Bashlykov, Batra, Bhargava, Bhosale, et~al.]{touvron2023llama}
Hugo Touvron, Louis Martin, Kevin Stone, Peter Albert, Amjad Almahairi, Yasmine Babaei, Nikolay Bashlykov, Soumya Batra, Prajjwal Bhargava, Shruti Bhosale, et~al.
\newblock Llama 2: Open foundation and fine-tuned chat models.
\newblock \emph{arXiv preprint arXiv:2307.09288}, 2023.

\bibitem[Wang et~al.(2018)Wang, Singh, Michael, Hill, Levy, and Bowman]{wang2018glue}
Alex Wang, Amanpreet Singh, Julian Michael, Felix Hill, Omer Levy, and Samuel~R Bowman.
\newblock {GLUE}: A multi-task benchmark and analysis platform for natural language understanding.
\newblock In \emph{Proceedings of the 2018 EMNLP workshop BlackboxNLP: Analyzing and interpreting neural networks for NLP}, pages 353--355, 2018.

\bibitem[Wei et~al.(2025)Wei, He, Yang, Liu, Wang, Feng, and An]{wei2025probsurgery}
Qi~Wei, Shuo He, Enneng Yang, Tingcong Liu, Haobo Wang, Lei Feng, and Bo~An.
\newblock Representation surgery in model merging with probabilistic modeling.
\newblock In \emph{Forty-second International Conference on Machine Learning}, 2025.

\bibitem[Xiao et~al.(2010)Xiao, Hays, Ehinger, Oliva, and Torralba]{xiao2010sun}
Jianxiong Xiao, James Hays, Krista~A Ehinger, Aude Oliva, and Antonio Torralba.
\newblock {SUN} database: Large-scale scene recognition from abbey to zoo.
\newblock In \emph{2010 IEEE computer society conference on computer vision and pattern recognition}, pages 3485--3492. IEEE, 2010.

\bibitem[Xu et~al.(2024)Xu, Sun, Zheng, Geng, Zhao, Feng, Tao, Lin, and Jiang]{xu2024wizardlm}
Can Xu, Qingfeng Sun, Kai Zheng, Xiubo Geng, Pu~Zhao, Jiazhan Feng, Chongyang Tao, Qingwei Lin, and Daxin Jiang.
\newblock {WizardLM}: Empowering large pre-trained language models to follow complex instructions.
\newblock In \emph{International Conference on Learning Representations}, pages 30745--30766, 2024.

\bibitem[Xu et~al.(2025)Xu, Li, and Zhang]{xu2025prodistill}
Jing Xu, Jiazheng Li, and Jingzhao Zhang.
\newblock Scalable model merging with progressive layer-wise distillation.
\newblock In \emph{Forty-second International Conference on Machine Learning}, 2025.
\newblock URL \url{https://openreview.net/forum?id=xX8NJShgny}.

\bibitem[Yadav et~al.(2023)Yadav, Tam, Choshen, Raffel, and Bansal]{yadav2023ties}
Prateek Yadav, Derek Tam, Leshem Choshen, Colin~A Raffel, and Mohit Bansal.
\newblock {TIES-Merging}: Resolving interference when merging models.
\newblock \emph{Advances in neural information processing systems}, 36:\penalty0 7093--7115, 2023.

\bibitem[Yang et~al.(2024{\natexlab{a}})Yang, Shen, Wang, Guo, Chen, Wang, and Tao]{yang2024surgery}
Enneng Yang, Li~Shen, Zhenyi Wang, Guibing Guo, Xiaojun Chen, Xingwei Wang, and Dacheng Tao.
\newblock Representation surgery for multi-task model merging.
\newblock In Ruslan Salakhutdinov, Zico Kolter, Katherine Heller, Adrian Weller, Nuria Oliver, Jonathan Scarlett, and Felix Berkenkamp, editors, \emph{Proceedings of the 41st International Conference on Machine Learning}, volume 235 of \emph{Proceedings of Machine Learning Research}, pages 56332--56356. PMLR, 21--27 Jul 2024{\natexlab{a}}.
\newblock URL \url{https://proceedings.mlr.press/v235/yang24t.html}.

\bibitem[Yang et~al.(2024{\natexlab{b}})Yang, Shen, Wang, Guo, Wang, Cao, Zhang, and Tao]{yang2024surgeryv2}
Enneng Yang, Li~Shen, Zhenyi Wang, Guibing Guo, Xingwei Wang, Xiaocun Cao, Jie Zhang, and Dacheng Tao.
\newblock {SurgeryV2}: Bridging the gap between model merging and multi-task learning with deep representation surgery.
\newblock \emph{arXiv preprint arXiv:2410.14389}, 2024{\natexlab{b}}.

\bibitem[Yang et~al.(2024{\natexlab{c}})Yang, Wang, Shen, Liu, Guo, Wang, and Tao]{yang2024adamerging}
Enneng Yang, Zhenyi Wang, Li~Shen, Shiwei Liu, Guibing Guo, Xingwei Wang, and Dacheng Tao.
\newblock {AdaMerging}: Adaptive model merging for multi-task learning.
\newblock In \emph{International Conference on Learning Representations}, pages 22743--22763, 2024{\natexlab{c}}.

\bibitem[Yang et~al.(2026)Yang, Shen, Guo, Wang, Cao, Zhang, and Tao]{yang2026ModelMergingSurvey}
Enneng Yang, Li~Shen, Guibing Guo, Xingwei Wang, Xiaochun Cao, Jie Zhang, and Dacheng Tao.
\newblock Model merging in {LLMs}, {MLLMs}, and beyond: Methods, theories, applications, and opportunities.
\newblock \emph{ACM Computing Surveys}, 58\penalty0 (8), February 2026.
\newblock ISSN 0360-0300.
\newblock \doi{10.1145/3787849}.
\newblock URL \url{https://doi.org/10.1145/3787849}.

\end{thebibliography}
\bibliographystyle{plainnat}

\clearpage
\begingroup
\hypersetup{linkcolor=black}  %
\tableofcontents
\endgroup
\clearpage
\appendix
\FloatBarrier
\section{Additional Related Work on Model Merging}
\label{app:additional_related_work}

Several model-merging approaches study the spectral structure of task updates to reduce interference and preserve task-specific information.
\method{TSV-Merge}~\citep{gargiulo2025tsv} decomposes layer-wise task matrices into singular components, retains their dominant directions, and orthogonalizes these singular vectors across tasks before reconstructing the merged update.
This combines low-rank compression with interference reduction.
\method{Iso-Merging}~\citep{marczak2025isomerging} studies the alignment between task-specific and merged updates and proposes flattening the singular value spectrum to balance the contributions of different directions.
Its framework also incorporates common and task-specific subspaces to preserve information that may be underrepresented in the aggregate update.
\method{Subspace Boosting}~\citep{skorobogat2025subspace} examines rank collapse as more experts are merged and boosts smaller singular values of the merged task update to retain a broader set of task-relevant directions.
Alignment between task updates also matters when merging models fine-tuned with low-rank adaptation.
\method{KnOTS}~\citep{stoica2025knots} observes that LoRA task updates exhibit less alignment than their fully fine-tuned counterparts, which can limit the effectiveness of existing merging methods.
It uses SVD to jointly transform the task updates into a common aligned space, applies an existing merging method in that space, and reconstructs the merged update.

\section{Motivation for Error-Subspace Calibration}
\label{app:error_subspace_motivation}

We examine the discrepancy between a task-specific model and the merged model across consecutive transformations.
We consider two properties of the merging error: the subspace containing the locally induced discrepancy at the current transformation, and the input directions relevant to the merging error of the following transformation.
Here, the transformations are the selected linear maps used for the error decompositions, not entire nonlinear Transformer blocks. We use a row-vector convention in which a linear transformation maps $\vx$ to $\vx\mW$, with all weight matrices written in input-by-output form.

We assume that the expert and merged model share the same architecture, with potentially different weights.
We fix a task index $t\in[T]$ and omit it when clear from context. 

\paragraph{Local merging error lies in the output error subspace.}

For a block $\ell\in [L]$, choose the same linear layer and take its weight matrix: $\mW_\ell^t$ for the expert and $\bar{\mW}_\ell$ for the merged model.
Recall that the corresponding merging error is $\mE_\ell^t = \mW_\ell^t - \bar{\mW}_\ell$.
Write its compact singular value decomposition (SVD) as $\mE_\ell^t = \mU_\ell^t \mSigma_\ell^t \mV_\ell^{t\top}$, where $\mSigma_\ell^t$ contains all nonzero singular values in descending order.
Under our row-vector convention, the columns of $\mU_\ell^t$ and $\mV_\ell^t$ span the input-side and output-side error subspaces of this transformation, respectively.

For a shared input representation $\vx$, the discrepancy between the task-specific and merged transformations is
\begin{equation}
\begin{aligned}
    \vx\mW_\ell^t - \vx\bar{\mW}_\ell
    &=
    \vx\mE_\ell^t \\
    &=
    \left(\vx\mU_\ell^t\right)
    \mSigma_\ell^t
    \mV_\ell^{t\top}.
\end{aligned}
\end{equation}
Therefore,
\begin{equation}
    \left(\vx\mE_\ell^t\right)^\top
    \in
    \operatorname{span}(\mV_\ell^t)
\end{equation}
for any shared input $\vx$. Thus, the discrepancy directly introduced by the
weight difference at transformation $\ell$ is confined to the output singular
subspace of the merging error.

This observation also implies that a correction outside this subspace cannot reduce the locally induced discrepancy.
Let a correction be decomposed as
\begin{equation}
    \Delta \vh_\ell
    =
    \Delta \vh_{\ell,\parallel}
    +
    \Delta \vh_{\ell,\perp},
\end{equation}
where
\begin{equation}
    \Delta \vh_{\ell,\parallel}^{\top}
    \in
    \operatorname{span}(\mV_\ell^t),
    \qquad
    \Delta \vh_{\ell,\perp}^{\top}
    \perp
    \operatorname{span}(\mV_\ell^t).
\end{equation}
Since $\left(\vx\mE_\ell^t\right)^\top$ lies in
$\operatorname{span}(\mV_\ell^t)$,
\begin{equation}
\begin{aligned}
    \left\|
        \vx\mE_\ell^t - \Delta \vh_\ell
    \right\|_2^2
    &=
    \left\|
        \vx\mE_\ell^t - \Delta \vh_{\ell,\parallel}
    \right\|_2^2
    +
    \left\|
        \Delta \vh_{\ell,\perp}
    \right\|_2^2.
\end{aligned}
\end{equation}
Hence, the orthogonal component $\Delta \vh_{\ell,\perp}$ cannot improve the
correction of the discrepancy directly induced by $\mE_\ell^t$. This motivates restricting the correction to
$\operatorname{span}(\mV_\ell^t)$.

This argument concerns the local discrepancy induced by the weight
difference under a shared input. The total representation discrepancy may
also contain errors propagated from preceding layers.

\paragraph{The following transformation identifies error-relevant input directions.}
We next consider the merging error of the transformation following the
calibration location. For an input representation $\vh$,
\begin{equation}
    \vh\mE_{\ell+1}^t
    =
    \left(\vh\mU_{\ell+1}^t\right)
    \mSigma_{\ell+1}^t
    \mV_{\ell+1}^{t\top}.
\end{equation}
The discrepancy induced by the following transformation therefore depends on
$\vh$ through its coordinates along $\mU_{\ell+1}^t$.

More specifically, let $\vu_{\ell+1,i}^t$ and $\vv_{\ell+1,i}^t$ denote the
$i$-th left and right singular vectors of $\mE_{\ell+1}^t$. Then
\begin{equation}
    \vu_{\ell+1,i}^{t\top}
    \mE_{\ell+1}^t
    =
    \sigma_{\ell+1,i}^t
    \vv_{\ell+1,i}^{t\top},
\end{equation}
and, since $\vv_{\ell+1,i}^t$ has unit norm,
\begin{equation}
    \left\|
        \vu_{\ell+1,i}^{t\top}
        \mE_{\ell+1}^t
    \right\|_2
    =
    \sigma_{\ell+1,i}^t.
\end{equation}
Thus, $\sigma_{\ell+1,i}^t$ measures the magnitude of the discrepancy
induced by the error operator for a unit input along
$\vu_{\ell+1,i}^t$.
Larger singular values therefore identify input directions along which
the following transformation's weight difference can induce larger local discrepancies.
This motivates using $\mU_{\ell+1}^t$ as the read subspace of \Bridge{}.

\paragraph{Scope of the argument.}
The quantity $\vh\mE_{\ell+1}^t$ is a local discrepancy caused by the
following transformation's weight difference. It is distinct from the
propagation of an existing representation perturbation $\Delta \vh$ through
the merged linear transformation, which is $\Delta \vh\bar{\mW}_{\ell+1}$.
For a nonlinear block, such propagation depends on the block's local
Jacobian.
Thus, the argument motivates the read and write subspaces of \Bridge{},
but does not imply optimality with respect to propagated error or the
total discrepancy of the nonlinear network.

\paragraph{Why retain the leading rank-$r$ subspaces?}
The arguments above concern the error subspaces spanned by all singular vectors associated with nonzero singular values. In practice, we
retain only the leading $r$ directions. For any transformation $\ell$, the rank-$r$
truncated SVD satisfies
\begin{equation}
    \left\|
        \mE_\ell^t
        -
        \mU_{\ell,r}^t
        \mSigma_{\ell,r}^t
        \mV_{\ell,r}^{t\top}
    \right\|_F^2
    =
    \sum_{i>r}
    \left(\sigma_{\ell,i}^t\right)^2.
\end{equation}
Truncated SVD minimizes this residual among rank-$r$ approximations of the
merging-error operator. For a fixed rank budget, it retains the dominant
merging-error directions and minimizes the discarded error energy in the Frobenius norm. This optimality concerns approximation of
the error matrix, not minimization of a data-weighted feature loss or
end-to-end prediction error.

These observations motivate using
$\mV_{\ell,r}^t$ as the output space for the correction at layer $\ell$ and
$\mU_{\ell+1,r}^t$ as the input space from which the correction is determined.
Accordingly, our \Bridge{} module is parameterized as
\begin{equation}
    \mB_\ell^t
    =
    \mU_{\ell+1,r}^t
    \mC_\ell^t
    \mV_{\ell,r}^{t\top},
    \qquad
    \mC_\ell^t \in \mathbb{R}^{r\times r}.
\end{equation}

The error subspaces are fixed by the task-specific and merged weights, while
calibration learns only the mapping $\mC_\ell^t$ between them.

The resulting feature correction can then be written as
\begin{equation}
    \Delta \vh_\ell^t
    =
    \vh_\ell \mB_\ell^t
    =
    \left(\vh_\ell \mU_{\ell+1,r}^t\right)
    \mC_\ell^t
    \mV_{\ell,r}^{t\top}.
\end{equation}
Hence, \Bridge{} reads coordinates from the following transformation's
input error subspace, maps them through $\mC_\ell^t$, and writes the resulting
correction into the current transformation's output error subspace.
\clearpage
\section{Experimental Details and Additional Results}
\label{app:additional_results}

We report results for additional model architectures, base merging methods,
and NLG task combinations, along with full per-task results and
variation across random seeds.
Unless otherwise noted, post-hoc calibration results are reported as the
mean $\pm$ sample standard deviation over five random seeds.
For stochastic base merging methods, we also report five-seed statistics
where available. Results from a single evaluation omit standard deviations.

\subsection{Experimental Settings}
\label{app:experimental_settings}

\paragraph{Vision.}
We consider three CLIP vision transformer architectures,
ViT-B/32, ViT-B/16, and ViT-L/14~\citep{radford2021clipvit}.
For each architecture, we use task-specific models on eight image
classification datasets:
SUN397~\citep{xiao2010sun},
Cars~\citep{krause2013cars},
RESISC45~\citep{cheng2017resics},
EuroSAT~\citep{helber2019eurosat},
SVHN~\citep{netzer2011svhn},
GTSRB~\citep{stallkamp2011gtsrb},
MNIST~\citep{lecun1998mnist},
and DTD~\citep{cimpoi2014dtd}.
The eight task-specific models are merged within each architecture,
and performance is evaluated on the corresponding test sets.
We report task-wise classification accuracy and its average across
the eight tasks.

\paragraph{NLU.}
We use RoBERTa-base~\citep{liu2019roberta} as the shared pretrained
model and consider eight tasks from the GLUE benchmark~\citep{wang2018glue}:
CoLA, SST-2, MRPC, STS-B, QQP, MNLI, QNLI, and RTE.
We merge the eight task-specific models and evaluate the resulting model
using the corresponding task-specific metrics.
The reported average is the arithmetic mean of the eight task scores.

\paragraph{NLG.}
We use LLaMA-2-13B~\citep{touvron2023llama} as the common base model
and consider task experts specialized for instruction following,
mathematical reasoning, and code generation.
The instruction-following and mathematical-reasoning experts are based on
WizardLM~\citep{xu2024wizardlm} and
WizardMath~\citep{luo2025wizardmath}, respectively.
We evaluate all three pairwise combinations of the task experts.
Detailed evaluation settings for the NLG experiments are provided
together with the corresponding results.

\subsection{Hyperparameter Search}
\label{app:hyperparameter_search}

We select hyperparameters independently for each base merging method
and benchmark setting.
Here, $r$ denotes the calibration rank, $\eta$ the learning rate,
and $E$ the number of training epochs.
We choose method-specific rank ranges with reference to trainable parameter
budgets. In the selected configurations, \ESCal{} uses more trainable
parameters than \Surgery{} and fewer than \Surgery{V2}.

\paragraph{Vision.}
For \Surgery{} and \Surgery{V2}, we fix the calibration rank at $r=16$.
Then we search the learning rate $\eta \in \{\text{1e-4}, \text{3e-4}, \text{1e-3}\}$
and the number of training epochs $E \in \{50, 100, 200, 250, 500\}$.
The search extends to 500 epochs (1,000 outer iterations under our vision calibration setup),
while the selected best setting for accuracy is 250 epochs (500 outer iterations).
The separate timing comparison in Table~\ref{tab:wallclock_cost} uses 200 epochs.
For \FeatCal{}, across all vision architectures and base merging methods,
we fix $\lambda=0.05$ and search
$\rho \in \{1, 1.5, 2, 2.5\}$ and
$\alpha \in \{0.1, 0.3, 0.5\}$.
For \ESCal{}, we fix $r=128$ and search
$\eta \in \{\text{1e-4}, \text{3e-4}, \text{1e-3}\}$ and
$E \in \{50, 100, 200\}$.

\paragraph{NLU.}
For \Surgery{}, \Surgery{V2}, and \ESCal{}, we search
$E \in \{200, 500, 1000\}$.
For \Surgery{} and \Surgery{V2}, we search $r \in \{4, 16, 64\}$ and
$\eta \in \{\text{1e-5}, \text{3e-5}, \text{1e-4}, \text{3e-4},
\text{1e-3}, \text{3e-3}, \text{1e-2}\}$.
For \FeatCal{}, we search
$\lambda \in \{\text{1e-3}, \text{3e-3}, \text{1e-2}, 0.05, 0.2\}$,
$\rho \in \{1, 1.5, 2, 2.5\}$, and
$\alpha \in \{0.1, 0.3, 0.5, 0.7, 0.9, 1.0\}$.
For \ESCal{}, we search
$r \in \{32, 64, 128\}$ and
$\eta \in \{\text{1e-4}, \text{3e-4}, \text{1e-3},
\text{3e-3}, \text{6e-3}, \text{1e-2}\}$.

\paragraph{NLG.}
For \Surgery{}, we search
$r \in \{16, 32\}$,
$\eta \in \{\text{1e-6}, \text{1e-5}, \text{1e-4}, \text{1e-3}\}$, and
$E \in \{5, 25, 50\}$.
For \Surgery{V2}, we search
$r \in \{8, 16\}$,
$\eta \in \{\text{1e-5}, \text{1e-4}, \text{1e-3}\}$, and
$E \in \{5, 25, 50\}$.
Because NLG evaluation is computationally expensive, we use a smaller
hyperparameter grid for \FeatCal{}.
We fix $\rho=1$ and search
$\lambda \in \{\text{1e-3}, \text{1e-1}\}$ and
$\alpha \in \{0.1, 0.3, 0.5\}$.
For \ESCal{}, we search
$r \in \{64, 128, 256\}$,
$\eta \in \{\text{1e-6}, \text{1e-5}, \text{1e-4},
\text{1e-3}\}$, and
$E \in \{5, 25, 50\}$.

\paragraph{IntervMerge.}
For \IntervMerge{}, we select a configuration independently for each architecture and base merging method using the existing seed-0 search results.
For vision, we search ranks $r\in\{1,4,16\}$, learning rates $\eta\in\{10^{-4},3\times10^{-4},10^{-3}\}$, and outer-iteration counts $K\in\{100,200,400,500,1000\}$.
The selection score is the arithmetic mean of the eight task test accuracies on seed 0; thus, the test sets are also used for configuration selection.
For NLU, we search $r\in\{1,4,16\}$, $\eta\in\{10^{-4},3\times10^{-4},10^{-3},3\times10^{-3},10^{-2}\}$, and $E\in\{100,200,500,1000\}$ epochs, selecting by the eight-task mean GLUE validation score on seed 0.
After selection, each configuration is fixed for seeds 0--4. We reuse the matching seed-0 search result and report the mean and sample standard deviation over these five runs; seeds 1--4 are not used for configuration selection.
Both settings use 64 calibration examples per task, batch size 16, Adam with constant learning rate, no warmup, and zero weight decay. The merged backbone remains frozen while task-specific interventions on the attention-output CLS representation are trained by final-representation L1 matching.
Vision uses two batches per task per outer iteration; NLU uses full passes over the task-specific calibration sets per epoch.
\begin{table}[htbp]
\centering\small
\caption{Selected IntervMerge configurations. Each vision entry is $(r,\eta,K)$, with $K$ measured in outer iterations.}
\label{tab:intervmerge_vision_settings}
\begin{tabular}{lccc}
\toprule
Base merging method & ViT-B/32 & ViT-B/16 & ViT-L/14 \\
\midrule
\TA{} & $(16,3\!\times\!10^{-4},1000)$ & $(16,10^{-3},200)$ & $(4,10^{-3},400)$ \\
\TIES{} & $(16,3\!\times\!10^{-4},500)$ & $(16,10^{-3},1000)$ & $(16,3\!\times\!10^{-4},1000)$ \\
\AdaMerging{} (tw) & $(16,10^{-3},400)$ & $(16,10^{-3},400)$ & $(4,3\!\times\!10^{-4},1000)$ \\
\AdaMerging{} (lw) & $(16,3\!\times\!10^{-4},400)$ & $(16,10^{-3},100)$ & $(16,3\!\times\!10^{-4},100)$ \\
\WUDI{} & $(16,10^{-4},500)$ & $(16,10^{-3},100)$ & $(16,3\!\times\!10^{-4},1000)$ \\
\ProDistill{} & $(16,10^{-3},1000)$ & $(16,3\!\times\!10^{-4},1000)$ & $(16,3\!\times\!10^{-4},500)$ \\
\bottomrule
\end{tabular}
\end{table}
\begin{table}[htbp]
\centering\small
\caption{Selected IntervMerge configurations for RoBERTa-base.}
\label{tab:intervmerge_nlu_settings}
\begin{tabular}{lccc}
\toprule
Base merging method & Rank $r$ & Learning rate $\eta$ & Epochs $E$ \\
\midrule
\TA{} & 4 & $3\times10^{-3}$ & 500 \\
\TIES{} & 16 & $10^{-3}$ & 200 \\
\AdaMerging{} (tw) & 16 & $3\times10^{-3}$ & 500 \\
\AdaMerging{} (lw) & 4 & $3\times10^{-3}$ & 1000 \\
\WUDI{} & 16 & $10^{-3}$ & 1000 \\
\ProDistill{} & 16 & $3\times10^{-3}$ & 1000 \\
\bottomrule
\end{tabular}
\end{table}

\subsection{Vision Results}
\label{app:vision_results}

Tables~\ref{tab:vision_b32_full}, \ref{tab:vision_b16_full},
and~\ref{tab:vision_l14_full} report the complete task-wise test
accuracies for ViT-B/32, ViT-B/16, and ViT-L/14, respectively,
across all six base merging methods and five post-merging calibration methods.
The \textit{Avg.} column reports the mean accuracy across the eight tasks.

\ESCal{} consistently improves over the corresponding uncalibrated models and achieves the highest average accuracy among the compared calibration methods in most settings.
\FeatCal{} performs best for \ProDistill{} across all three architectures.

\begin{table}[htbp]
\centering
\renewcommand{\arraystretch}{1.1}
\caption{Full task-wise test accuracy (\%) on ViT-B/32.}
\label{tab:vision_b32_full}
\resizebox{\textwidth}{!}{
\begin{tabular}{lcccccccc|c}
\toprule
\textbf{Method}
& \textbf{SUN397}
& \textbf{Cars}
& \textbf{RESISC45}
& \textbf{EuroSAT}
& \textbf{SVHN}
& \textbf{GTSRB}
& \textbf{MNIST}
& \textbf{DTD}
& \textbf{Avg.} \\
\midrule

\TA{}
& 55.32 & 54.98 & 66.82 & 77.78 & 80.21 & 69.68 & 97.34 & 50.37 & 69.06 \\
\quad w/ \Surgery{}
& 60.39 $\pm$ 0.46 & 52.62 $\pm$ 0.59 & 71.38 $\pm$ 1.76
& 88.65 $\pm$ 0.79 & 81.67 $\pm$ 1.65 & 74.27 $\pm$ 0.91
& 97.82 $\pm$ 0.16 & 52.77 $\pm$ 2.73 & 72.45 $\pm$ 0.52 \\
\quad w/ \Surgery{V2}
& \textbf{70.70} $\pm$ 0.19 & 66.41 $\pm$ 0.49 & 90.65 $\pm$ 1.23
& 99.12 $\pm$ 0.18 & 96.61 $\pm$ 0.07 & 95.88 $\pm$ 0.62
& 99.47 $\pm$ 0.06 & 70.90 $\pm$ 0.75 & 86.22 $\pm$ 0.17 \\
\quad w/ \IntervMerge{}
& 66.35 $\pm$ 0.38 & 58.87 $\pm$ 0.83 & 85.88 $\pm$ 1.55 & 97.61 $\pm$ 0.32 & 93.44 $\pm$ 0.22 & 87.49 $\pm$ 1.09 & 98.62 $\pm$ 0.10 & 64.31 $\pm$ 1.10 & 81.57 $\pm$ 0.29 \\
\quad w/ \FeatCal{}
& 69.22 & \textbf{70.31} & 89.06 & 98.00 & 95.01 & 92.40
& 99.13 & 71.11 & 85.53 \\
\quad \textbf{w/ \ESCal{}}
& 70.33 $\pm$ 0.22 & 69.02 $\pm$ 0.19 & \textbf{93.60} $\pm$ 0.36
& \textbf{99.48} $\pm$ 0.07 & \textbf{97.10} $\pm$ 0.05
& \textbf{97.70} $\pm$ 0.22 & \textbf{99.55} $\pm$ 0.03
& \textbf{73.52} $\pm$ 0.44 & \textbf{87.54} $\pm$ 0.07 \\

\midrule

\TIES{}
& 65.05 & 64.43 & 74.92 & 76.48 & 81.28 & 69.37 & 96.53 & 54.52 & 72.82 \\
\quad w/ \Surgery{}
& 67.64 $\pm$ 0.39 & 59.57 $\pm$ 0.54 & 78.50 $\pm$ 1.77
& 89.24 $\pm$ 1.30 & 81.30 $\pm$ 1.43 & 76.07 $\pm$ 0.29
& 97.14 $\pm$ 0.31 & 60.03 $\pm$ 1.81 & 76.19 $\pm$ 0.39 \\
\quad w/ \Surgery{V2}
& 70.87 $\pm$ 0.20 & 66.26 $\pm$ 0.75 & 90.14 $\pm$ 1.71
& 98.94 $\pm$ 0.21 & 96.54 $\pm$ 0.09 & 95.52 $\pm$ 0.77
& 99.45 $\pm$ 0.07 & 70.89 $\pm$ 0.72 & 86.08 $\pm$ 0.21 \\
\quad w/ \IntervMerge{}
& 68.65 $\pm$ 0.28 & 60.59 $\pm$ 1.12 & 86.21 $\pm$ 1.75 & 97.39 $\pm$ 0.52 & 92.99 $\pm$ 0.23 & 87.60 $\pm$ 0.97 & 98.24 $\pm$ 0.18 & 65.14 $\pm$ 1.17 & 82.10 $\pm$ 0.26 \\
\quad w/ \FeatCal{}
& 70.12 $\pm$ 0.02 & \textbf{70.74} $\pm$ 0.17 & 89.28 $\pm$ 0.31
& 97.93 $\pm$ 0.13 & 94.71 $\pm$ 0.02 & 92.55 $\pm$ 0.28
& 99.10 $\pm$ 0.03 & 71.13 $\pm$ 0.70 & 85.69 $\pm$ 0.13 \\
\quad \textbf{w/ \ESCal{}}
& \textbf{70.93} $\pm$ 0.23 & 69.23 $\pm$ 0.18
& \textbf{93.31} $\pm$ 0.48 & \textbf{99.36} $\pm$ 0.10
& \textbf{97.09} $\pm$ 0.03 & \textbf{97.70} $\pm$ 0.25
& \textbf{99.53} $\pm$ 0.04 & \textbf{73.35} $\pm$ 0.43
& \textbf{87.56} $\pm$ 0.07 \\

\midrule

\AdaMerging{} (tw)
& 58.80 $\pm$ 3.16 & 53.42 $\pm$ 1.76 & 64.91 $\pm$ 6.63
& 86.51 $\pm$ 1.86 & 79.67 $\pm$ 4.41 & 83.34 $\pm$ 4.13
& 90.32 $\pm$ 2.60 & 49.55 $\pm$ 8.56 & 70.82 $\pm$ 1.04 \\
\quad w/ \Surgery{}
& 61.90 $\pm$ 2.48 & 49.77 $\pm$ 2.25 & 69.97 $\pm$ 7.03
& 92.50 $\pm$ 1.03 & 80.29 $\pm$ 4.93 & 85.75 $\pm$ 4.32
& 94.73 $\pm$ 1.04 & 52.66 $\pm$ 9.39 & 73.45 $\pm$ 1.22 \\
\quad w/ \Surgery{V2}
& 69.86 $\pm$ 1.86 & 61.67 $\pm$ 2.43 & 88.89 $\pm$ 2.83
& 99.25 $\pm$ 0.13 & 96.37 $\pm$ 0.22 & 96.97 $\pm$ 0.82
& 99.32 $\pm$ 0.15 & 68.57 $\pm$ 5.36 & 85.11 $\pm$ 0.55 \\
\quad w/ \IntervMerge{}
& 66.08 $\pm$ 2.21 & 53.08 $\pm$ 3.70 & 82.23 $\pm$ 4.76 & 98.36 $\pm$ 0.37 & 92.45 $\pm$ 1.46 & 92.67 $\pm$ 2.45 & 96.77 $\pm$ 1.33 & 61.95 $\pm$ 7.19 & 80.45 $\pm$ 0.86 \\
\quad w/ \FeatCal{}
& 68.86 $\pm$ 2.21 & \textbf{66.32} $\pm$ 2.14 & 86.92 $\pm$ 2.34
& 98.42 $\pm$ 0.39 & 94.72 $\pm$ 0.58 & 94.68 $\pm$ 1.13
& 98.51 $\pm$ 0.37 & 69.31 $\pm$ 4.99 & 84.72 $\pm$ 0.52 \\
\quad \textbf{w/ \ESCal{}}
& \textbf{70.10} $\pm$ 1.65 & 65.55 $\pm$ 1.43
& \textbf{92.79} $\pm$ 1.32 & \textbf{99.59} $\pm$ 0.05
& \textbf{97.03} $\pm$ 0.12 & \textbf{98.06} $\pm$ 0.27
& \textbf{99.51} $\pm$ 0.05 & \textbf{72.35} $\pm$ 3.07
& \textbf{86.87} $\pm$ 0.29 \\

\midrule

\AdaMerging{} (lw)
& 63.28 $\pm$ 2.89 & 66.71 $\pm$ 2.16 & 79.15 $\pm$ 3.50
& 89.81 $\pm$ 1.31 & 83.82 $\pm$ 3.25 & 90.56 $\pm$ 3.91
& 97.87 $\pm$ 0.48 & 63.28 $\pm$ 3.76 & 79.31 $\pm$ 0.76 \\
\quad w/ \Surgery{}
& 66.33 $\pm$ 2.19 & 64.87 $\pm$ 2.88 & 84.24 $\pm$ 2.56
& 94.24 $\pm$ 0.46 & 87.64 $\pm$ 2.57 & 92.83 $\pm$ 3.24
& 98.30 $\pm$ 0.41 & 67.12 $\pm$ 3.61 & 81.95 $\pm$ 0.85 \\
\quad w/ \Surgery{V2}
& \textbf{71.45} $\pm$ 1.26 & 70.24 $\pm$ 2.27 & 92.79 $\pm$ 0.71
& 99.14 $\pm$ 0.07 & 96.54 $\pm$ 0.12 & 98.08 $\pm$ 0.31
& 99.54 $\pm$ 0.03 & \textbf{75.45} $\pm$ 1.56
& 87.90 $\pm$ 0.31 \\
\quad w/ \IntervMerge{}
& 68.94 $\pm$ 2.09 & 66.72 $\pm$ 2.74 & 89.75 $\pm$ 1.04 & 98.17 $\pm$ 0.31 & 92.83 $\pm$ 0.78 & 96.63 $\pm$ 0.82 & 98.83 $\pm$ 0.21 & 72.34 $\pm$ 2.78 & 85.53 $\pm$ 0.37 \\
\quad w/ \FeatCal{}
& 70.68 $\pm$ 1.70 & \textbf{74.08} $\pm$ 1.41 & 91.16 $\pm$ 0.59
& 98.05 $\pm$ 0.13 & 94.69 $\pm$ 0.27 & 97.03 $\pm$ 0.52
& 99.21 $\pm$ 0.09 & 75.06 $\pm$ 1.31 & 87.50 $\pm$ 0.31 \\
\quad \textbf{w/ \ESCal{}}
& 69.51 $\pm$ 1.89 & 70.99 $\pm$ 1.43 & \textbf{93.40} $\pm$ 0.36
& \textbf{99.48} $\pm$ 0.11 & \textbf{97.04} $\pm$ 0.06
& \textbf{98.38} $\pm$ 0.23 & \textbf{99.58} $\pm$ 0.03
& 74.87 $\pm$ 0.94 & \textbf{87.91} $\pm$ 0.24 \\

\midrule

\WUDI{}
& 69.11 & 71.38 & 85.62 & 95.63 & 94.27 & 94.51 & 99.47 & 69.47 & 84.93 \\
\quad w/ \Surgery{}
& 69.24 $\pm$ 0.22 & 69.42 $\pm$ 0.22 & 87.70 $\pm$ 0.20
& 97.53 $\pm$ 0.33 & 94.81 $\pm$ 0.07 & 95.43 $\pm$ 0.34
& 99.48 $\pm$ 0.04 & 71.37 $\pm$ 0.49 & 85.62 $\pm$ 0.17 \\
\quad w/ \Surgery{V2}
& 73.24 $\pm$ 0.08 & 73.57 $\pm$ 0.45 & 94.15 $\pm$ 0.13
& \textbf{99.59} $\pm$ 0.06 & 97.17 $\pm$ 0.07
& 98.38 $\pm$ 0.17 & \textbf{99.64} $\pm$ 0.01
& \textbf{77.66} $\pm$ 0.61 & 89.18 $\pm$ 0.09 \\
\quad w/ \IntervMerge{}
& 71.72 $\pm$ 0.25 & 72.55 $\pm$ 0.21 & 92.65 $\pm$ 0.15 & 99.21 $\pm$ 0.20 & 96.33 $\pm$ 0.07 & 97.97 $\pm$ 0.23 & 99.59 $\pm$ 0.03 & 75.89 $\pm$ 0.74 & 88.24 $\pm$ 0.11 \\
\quad w/ \FeatCal{}
& 71.98 $\pm$ 0.14 & \textbf{75.80} $\pm$ 0.17
& 93.71 $\pm$ 0.08 & 99.31 $\pm$ 0.07
& 96.53 $\pm$ 0.04 & 98.42 $\pm$ 0.03
& 99.50 $\pm$ 0.01 & 77.52 $\pm$ 0.24 & 89.10 $\pm$ 0.05 \\
\quad \textbf{w/ \ESCal{}}
& \textbf{73.55} $\pm$ 0.13 & 74.68 $\pm$ 0.15
& \textbf{94.80} $\pm$ 0.14 & 99.58 $\pm$ 0.04
& \textbf{97.31} $\pm$ 0.01 & \textbf{98.68} $\pm$ 0.07
& \textbf{99.64} $\pm$ 0.02 & 77.52 $\pm$ 0.40
& \textbf{89.47} $\pm$ 0.06 \\

\midrule

\ProDistill{}
& 68.84 $\pm$ 0.20 & 70.50 $\pm$ 0.29
& 90.74 $\pm$ 0.81 & 99.06 $\pm$ 0.14
& 96.35 $\pm$ 0.35 & 96.17 $\pm$ 0.63
& 99.50 $\pm$ 0.03 & 67.53 $\pm$ 0.74
& 86.09 $\pm$ 0.12 \\
\quad w/ \Surgery{}
& 69.55 $\pm$ 0.14 & \textbf{70.56} $\pm$ 0.42
& 90.90 $\pm$ 0.80 & 99.08 $\pm$ 0.12
& 96.33 $\pm$ 0.35 & 96.07 $\pm$ 0.68
& 99.51 $\pm$ 0.03 & 68.98 $\pm$ 0.79 & 86.37 $\pm$ 0.13 \\
\quad w/ \Surgery{V2}
& \textbf{70.09} $\pm$ 0.25 & 70.38 $\pm$ 0.35
& 91.21 $\pm$ 0.99 & 98.93 $\pm$ 0.16
& 96.85 $\pm$ 0.14 & 96.25 $\pm$ 0.67
& 99.48 $\pm$ 0.03 & 69.29 $\pm$ 0.88 & 86.56 $\pm$ 0.19 \\
\quad w/ \IntervMerge{}
& 69.48 $\pm$ 0.31 & 70.09 $\pm$ 0.37 & 90.97 $\pm$ 1.00 & 99.06 $\pm$ 0.14 & 96.34 $\pm$ 0.36 & 96.11 $\pm$ 0.64 & 99.51 $\pm$ 0.03 & 68.94 $\pm$ 0.86 & 86.31 $\pm$ 0.12 \\
\quad w/ \FeatCal{}
& 69.63 $\pm$ 0.14 & 69.80 $\pm$ 0.45
& \textbf{92.60} $\pm$ 0.31 & 99.08 $\pm$ 0.16
& \textbf{96.88} $\pm$ 0.11 & 96.35 $\pm$ 0.55
& 99.51 $\pm$ 0.04 & \textbf{72.44} $\pm$ 0.68
& \textbf{87.03} $\pm$ 0.13 \\
\quad \textbf{w/ \ESCal{}}
& 69.21 $\pm$ 0.26 & 70.44 $\pm$ 0.40
& 91.30 $\pm$ 0.87 & \textbf{99.19} $\pm$ 0.13
& 96.87 $\pm$ 0.11 & \textbf{96.89} $\pm$ 0.51
& \textbf{99.53} $\pm$ 0.03 & 68.59 $\pm$ 1.02
& 86.50 $\pm$ 0.14 \\

\bottomrule
\end{tabular}
}
\end{table}

\begin{table}[htbp]
\centering
\renewcommand{\arraystretch}{1.1}
\caption{Full task-wise test accuracy (\%) on ViT-B/16.}
\label{tab:vision_b16_full}
\resizebox{\textwidth}{!}{
\begin{tabular}{lcccccccc|c}
\toprule
\textbf{Method}
& \textbf{SUN397}
& \textbf{Cars}
& \textbf{RESISC45}
& \textbf{EuroSAT}
& \textbf{SVHN}
& \textbf{GTSRB}
& \textbf{MNIST}
& \textbf{DTD}
& \textbf{Avg.} \\
\midrule

\TA{}
& 61.74 & 65.96 & 74.15 & 77.07 & 88.06 & 73.92 & 98.48 & 53.03 & 74.05 \\
\quad w/ \Surgery{}
& 65.25 $\pm$ 0.43 & 63.49 $\pm$ 1.63 & 76.73 $\pm$ 1.36
& 89.67 $\pm$ 0.99 & 88.03 $\pm$ 0.40 & 76.13 $\pm$ 2.11
& 98.39 $\pm$ 0.12 & 58.62 $\pm$ 2.21 & 77.04 $\pm$ 0.40 \\
\quad w/ \Surgery{V2}
& 74.17 $\pm$ 0.16 & 76.68 $\pm$ 0.75 & 93.02 $\pm$ 0.59
& 98.95 $\pm$ 0.24 & 97.21 $\pm$ 0.06 & 96.51 $\pm$ 0.62
& 99.52 $\pm$ 0.04 & 74.85 $\pm$ 0.95 & 88.86 $\pm$ 0.17 \\
\quad w/ \IntervMerge{}
& 70.34 $\pm$ 0.22 & 69.55 $\pm$ 0.96 & 88.15 $\pm$ 2.12 & 97.50 $\pm$ 0.45 & 95.30 $\pm$ 0.22 & 86.57 $\pm$ 2.14 & 99.00 $\pm$ 0.12 & 67.57 $\pm$ 1.88 & 84.25 $\pm$ 0.36 \\
\quad w/ \FeatCal{}
& 71.96 $\pm$ 0.06 & \textbf{79.84} $\pm$ 0.19
& 90.12 $\pm$ 0.32 & 97.95 $\pm$ 0.26
& 96.44 $\pm$ 0.03 & 93.37 $\pm$ 0.03
& 99.16 $\pm$ 0.03 & 73.41 $\pm$ 0.65 & 87.78 $\pm$ 0.11 \\
\quad \textbf{w/ \ESCal{}}
& \textbf{74.22} $\pm$ 0.27 & 79.82 $\pm$ 0.54
& \textbf{95.07} $\pm$ 0.26 & \textbf{99.46} $\pm$ 0.09
& \textbf{97.61} $\pm$ 0.08 & \textbf{98.47} $\pm$ 0.26
& \textbf{99.58} $\pm$ 0.02 & \textbf{77.31} $\pm$ 0.63
& \textbf{90.19} $\pm$ 0.20 \\

\midrule

\TIES{}
& 69.02 & 72.63 & 80.70 & 84.56 & 85.08 & 71.56 & 98.19 & 54.95 & 77.09 \\
\quad w/ \Surgery{}
& 71.14 $\pm$ 0.22 & 68.61 $\pm$ 1.21 & 82.55 $\pm$ 2.36
& 92.36 $\pm$ 1.13 & 84.04 $\pm$ 1.31 & 75.18 $\pm$ 2.78
& 98.10 $\pm$ 0.26 & 63.81 $\pm$ 2.36 & 79.47 $\pm$ 0.40 \\
\quad w/ \Surgery{V2}
& 74.30 $\pm$ 0.09 & 75.93 $\pm$ 0.92 & 92.40 $\pm$ 1.02
& 98.70 $\pm$ 0.36 & 97.18 $\pm$ 0.09 & 96.22 $\pm$ 0.55
& 99.51 $\pm$ 0.02 & 74.89 $\pm$ 1.25 & 88.64 $\pm$ 0.20 \\
\quad w/ \IntervMerge{}
& 71.65 $\pm$ 0.23 & 69.77 $\pm$ 0.97 & 87.73 $\pm$ 2.36 & 97.22 $\pm$ 0.52 & 94.93 $\pm$ 0.34 & 85.45 $\pm$ 2.60 & 98.91 $\pm$ 0.14 & 67.83 $\pm$ 1.72 & 84.19 $\pm$ 0.49 \\
\quad w/ \FeatCal{}
& 73.67 $\pm$ 0.03 & \textbf{81.47} $\pm$ 0.10
& 90.71 $\pm$ 0.19 & 98.01 $\pm$ 0.14
& 96.38 $\pm$ 0.01 & 93.99 $\pm$ 0.09
& 99.13 $\pm$ 0.03 & 74.04 $\pm$ 0.46 & 88.42 $\pm$ 0.07 \\
\quad \textbf{w/ \ESCal{}}
& \textbf{74.50} $\pm$ 0.30 & 79.98 $\pm$ 0.32
& \textbf{95.13} $\pm$ 0.26 & \textbf{99.40} $\pm$ 0.09
& \textbf{97.53} $\pm$ 0.06 & \textbf{98.38} $\pm$ 0.28
& \textbf{99.57} $\pm$ 0.02 & \textbf{77.36} $\pm$ 0.86
& \textbf{90.23} $\pm$ 0.15 \\

\midrule

\AdaMerging{} (tw)
& 66.63 $\pm$ 2.45 & 62.80 $\pm$ 4.01 & 71.15 $\pm$ 5.51
& 84.53 $\pm$ 6.46 & 86.88 $\pm$ 4.56 & 91.13 $\pm$ 1.93
& 96.95 $\pm$ 0.57 & 42.65 $\pm$ 7.08 & 75.34 $\pm$ 0.71 \\
\quad w/ \Surgery{}
& 68.31 $\pm$ 2.16 & 60.42 $\pm$ 4.33 & 75.25 $\pm$ 5.05
& 91.07 $\pm$ 3.68 & 87.31 $\pm$ 5.25 & 90.51 $\pm$ 2.74
& 97.08 $\pm$ 0.46 & 49.45 $\pm$ 8.83 & 77.43 $\pm$ 1.00 \\
\quad w/ \Surgery{V2}
& 74.03 $\pm$ 1.56 & 72.72 $\pm$ 3.69 & 91.17 $\pm$ 2.14
& 98.97 $\pm$ 0.37 & 97.07 $\pm$ 0.22 & 97.98 $\pm$ 0.46
& 99.44 $\pm$ 0.07 & 67.66 $\pm$ 5.50 & 87.38 $\pm$ 0.47 \\
\quad w/ \IntervMerge{}
& 70.99 $\pm$ 1.90 & 64.34 $\pm$ 4.09 & 85.25 $\pm$ 4.33 & 97.66 $\pm$ 0.86 & 94.72 $\pm$ 1.64 & 94.41 $\pm$ 1.84 & 98.35 $\pm$ 0.37 & 58.23 $\pm$ 6.90 & 83.00 $\pm$ 0.95 \\
\quad w/ \FeatCal{}
& 72.91 $\pm$ 1.72 & 75.85 $\pm$ 3.47 & 87.93 $\pm$ 2.15
& 98.01 $\pm$ 0.63 & 96.27 $\pm$ 0.49 & 96.52 $\pm$ 0.57
& 98.87 $\pm$ 0.10 & 62.30 $\pm$ 7.94 & 86.08 $\pm$ 0.66 \\
\quad \textbf{w/ \ESCal{}}
& \textbf{74.84} $\pm$ 1.18 & \textbf{77.85} $\pm$ 2.18
& \textbf{94.78} $\pm$ 0.59 & \textbf{99.47} $\pm$ 0.19
& \textbf{97.50} $\pm$ 0.11 & \textbf{98.80} $\pm$ 0.12
& \textbf{99.51} $\pm$ 0.03 & \textbf{72.20} $\pm$ 3.90
& \textbf{89.37} $\pm$ 0.32 \\

\midrule

\AdaMerging{} (lw)
& 68.96 $\pm$ 1.85 & 80.02 $\pm$ 2.77 & 76.80 $\pm$ 6.10
& 86.30 $\pm$ 4.65 & 92.06 $\pm$ 0.63 & 94.35 $\pm$ 2.48
& 98.22 $\pm$ 0.59 & 62.54 $\pm$ 8.08 & 82.41 $\pm$ 1.08 \\
\quad w/ \Surgery{}
& 70.45 $\pm$ 1.38 & 78.87 $\pm$ 2.64 & 81.87 $\pm$ 3.01
& 91.79 $\pm$ 1.85 & 92.04 $\pm$ 0.84 & 94.88 $\pm$ 1.86
& 98.50 $\pm$ 0.47 & 68.65 $\pm$ 5.21 & 84.63 $\pm$ 0.71 \\
\quad w/ \Surgery{V2}
& \textbf{75.42} $\pm$ 1.08 & 83.29 $\pm$ 1.17
& 93.90 $\pm$ 1.17 & 99.06 $\pm$ 0.18
& 97.24 $\pm$ 0.17 & 98.70 $\pm$ 0.18
& 99.55 $\pm$ 0.05 & \textbf{77.53} $\pm$ 2.69
& 90.59 $\pm$ 0.26 \\
\quad w/ \IntervMerge{}
& 72.81 $\pm$ 1.46 & 80.42 $\pm$ 2.11 & 88.93 $\pm$ 2.72 & 97.85 $\pm$ 0.41 & 95.33 $\pm$ 0.56 & 96.95 $\pm$ 0.93 & 99.06 $\pm$ 0.16 & 72.43 $\pm$ 5.05 & 87.97 $\pm$ 0.39 \\
\quad w/ \FeatCal{}
& 72.58 $\pm$ 0.99 & \textbf{84.67} $\pm$ 1.29
& 89.75 $\pm$ 3.05 & 98.09 $\pm$ 0.15
& 96.20 $\pm$ 0.20 & 97.11 $\pm$ 0.30
& 99.11 $\pm$ 0.13 & 74.84 $\pm$ 3.85
& 89.04 $\pm$ 0.45 \\
\quad \textbf{w/ \ESCal{}}
& 74.66 $\pm$ 0.69 & 83.25 $\pm$ 0.87
& \textbf{95.06} $\pm$ 0.64 & \textbf{99.43} $\pm$ 0.09
& \textbf{97.58} $\pm$ 0.06 & \textbf{98.85} $\pm$ 0.10
& \textbf{99.57} $\pm$ 0.05 & 77.21 $\pm$ 2.40
& \textbf{90.70} $\pm$ 0.26 \\

\midrule

\WUDI{}
& 73.29 & 81.71 & 90.07 & 98.19 & 95.30 & 96.55 & 99.44 & 73.78 & 88.54 \\
\quad w/ \Surgery{}
& 73.68 $\pm$ 0.18 & 81.00 $\pm$ 0.36 & 91.34 $\pm$ 0.30
& 98.60 $\pm$ 0.13 & 95.09 $\pm$ 0.21 & 96.42 $\pm$ 0.25
& 99.45 $\pm$ 0.02 & 76.21 $\pm$ 0.85 & 88.97 $\pm$ 0.11 \\
\quad w/ \Surgery{V2}
& 76.68 $\pm$ 0.11 & 84.13 $\pm$ 0.41
& 95.70 $\pm$ 0.14 & \textbf{99.64} $\pm$ 0.08
& 97.67 $\pm$ 0.07 & 98.90 $\pm$ 0.10
& \textbf{99.67} $\pm$ 0.01 & 80.45 $\pm$ 0.70
& 91.60 $\pm$ 0.10 \\
\quad w/ \IntervMerge{}
& 75.01 $\pm$ 0.17 & 82.82 $\pm$ 0.44 & 94.18 $\pm$ 0.44 & 99.29 $\pm$ 0.10 & 97.13 $\pm$ 0.09 & 98.44 $\pm$ 0.28 & 99.57 $\pm$ 0.02 & 79.21 $\pm$ 0.52 & 90.71 $\pm$ 0.10 \\
\quad w/ \FeatCal{}
& 75.64 $\pm$ 0.10 & \textbf{84.62} $\pm$ 0.15
& 94.52 $\pm$ 0.10 & 98.96 $\pm$ 0.06
& 97.27 $\pm$ 0.03 & 98.46 $\pm$ 0.04
& 99.46 $\pm$ 0.02 & 79.50 $\pm$ 0.29
& 91.06 $\pm$ 0.02 \\
\quad \textbf{w/ \ESCal{}}
& \textbf{76.78} $\pm$ 0.06 & 84.28 $\pm$ 0.27
& \textbf{96.02} $\pm$ 0.05 & 99.61 $\pm$ 0.07
& \textbf{97.78} $\pm$ 0.04 & \textbf{98.95} $\pm$ 0.07
& 99.65 $\pm$ 0.02 & \textbf{80.50} $\pm$ 0.19
& \textbf{91.70} $\pm$ 0.05 \\

\midrule

\ProDistill{}
& 72.58 $\pm$ 0.08 & 81.01 $\pm$ 0.28 & 93.07 $\pm$ 0.86
& 99.19 $\pm$ 0.10 & 97.34 $\pm$ 0.16 & 97.65 $\pm$ 0.33
& 99.59 $\pm$ 0.02 & 71.35 $\pm$ 0.89 & 88.97 $\pm$ 0.17 \\
\quad w/ \Surgery{}
& 73.26 $\pm$ 0.22 & 81.12 $\pm$ 0.36 & 92.98 $\pm$ 0.94
& 99.21 $\pm$ 0.11 & 97.34 $\pm$ 0.15 & 97.41 $\pm$ 0.38
& 99.59 $\pm$ 0.03 & 72.66 $\pm$ 1.07 & 89.20 $\pm$ 0.18 \\
\quad w/ \Surgery{V2}
& \textbf{73.73} $\pm$ 0.18 & \textbf{81.32} $\pm$ 0.51
& 93.14 $\pm$ 0.98 & 99.24 $\pm$ 0.15
& 97.42 $\pm$ 0.10 & 97.47 $\pm$ 0.28
& 99.60 $\pm$ 0.02 & 73.30 $\pm$ 0.79
& 89.40 $\pm$ 0.19 \\
\quad w/ \IntervMerge{}
& 73.41 $\pm$ 0.17 & 80.84 $\pm$ 0.50 & 93.14 $\pm$ 0.97 & 99.19 $\pm$ 0.13 & 97.35 $\pm$ 0.16 & 97.46 $\pm$ 0.31 & 99.61 $\pm$ 0.03 & 72.80 $\pm$ 0.76 & 89.23 $\pm$ 0.18 \\
\quad w/ \FeatCal{}
& 72.26 $\pm$ 0.35 & 80.29 $\pm$ 0.23
& \textbf{94.65} $\pm$ 0.39 & \textbf{99.27} $\pm$ 0.06
& \textbf{97.53} $\pm$ 0.10 & 97.93 $\pm$ 0.22
& \textbf{99.68} $\pm$ 0.01 & \textbf{76.56} $\pm$ 0.50
& \textbf{89.77} $\pm$ 0.05 \\
\quad \textbf{w/ \ESCal{}}
& 72.96 $\pm$ 0.02 & 81.02 $\pm$ 0.37
& 93.52 $\pm$ 0.70 & 99.19 $\pm$ 0.16
& 97.50 $\pm$ 0.13 & \textbf{97.95} $\pm$ 0.27
& 99.60 $\pm$ 0.01 & 72.47 $\pm$ 1.04
& 89.27 $\pm$ 0.17 \\

\bottomrule
\end{tabular}
}
\end{table}

\begin{table}[htbp]
\centering
\renewcommand{\arraystretch}{1.1}

\caption{Full task-wise test accuracy (\%) on ViT-L/14.}
\label{tab:vision_l14_full}
\resizebox{\textwidth}{!}{
\begin{tabular}{lcccccccc|c}
\toprule
\textbf{Method}
& \textbf{SUN397}
& \textbf{Cars}
& \textbf{RESISC45}
& \textbf{EuroSAT}
& \textbf{SVHN}
& \textbf{GTSRB}
& \textbf{MNIST}
& \textbf{DTD}
& \textbf{Avg.} \\
\midrule

\TA{}
& 74.16 & 82.09 & 86.67 & 93.93 & 87.91 & 86.77 & 98.94 & 65.69 & 84.52 \\
\quad w/ \Surgery{}
& 74.71 $\pm$ 0.27 & 80.50 $\pm$ 0.43 & 87.89 $\pm$ 1.66
& 95.47 $\pm$ 0.57 & 88.82 $\pm$ 0.59 & 86.87 $\pm$ 1.01
& 98.94 $\pm$ 0.13 & 69.84 $\pm$ 1.94 & 85.38 $\pm$ 0.19 \\
\quad w/ \Surgery{V2}
& 78.61 $\pm$ 0.28 & 86.93 $\pm$ 0.31 & 94.57 $\pm$ 1.00
& 99.39 $\pm$ 0.04 & 97.73 $\pm$ 0.04 & 97.55 $\pm$ 0.54
& 99.58 $\pm$ 0.03 & 78.62 $\pm$ 0.99 & 91.62 $\pm$ 0.11 \\
\quad w/ \IntervMerge{}
& 76.40 $\pm$ 0.26 & 83.59 $\pm$ 0.68 & 91.62 $\pm$ 1.51 & 98.53 $\pm$ 0.30 & 96.29 $\pm$ 0.15 & 92.54 $\pm$ 1.61 & 99.42 $\pm$ 0.04 & 74.19 $\pm$ 1.01 & 89.07 $\pm$ 0.10 \\
\quad w/ \FeatCal{}
& 79.05 $\pm$ 0.04 & \textbf{90.18} $\pm$ 0.09
& 94.11 $\pm$ 0.11 & 99.30 $\pm$ 0.05
& 96.47 $\pm$ 0.04 & 96.85 $\pm$ 0.08
& 99.36 $\pm$ 0.03 & 78.96 $\pm$ 0.06
& 91.78 $\pm$ 0.02 \\
\quad \textbf{w/ \ESCal{}}
& \textbf{79.12} $\pm$ 0.17 & 89.28 $\pm$ 0.16
& \textbf{96.11} $\pm$ 0.28 & \textbf{99.66} $\pm$ 0.05
& \textbf{97.98} $\pm$ 0.03 & \textbf{98.75} $\pm$ 0.23
& \textbf{99.64} $\pm$ 0.02 & \textbf{80.52} $\pm$ 0.28
& \textbf{92.63} $\pm$ 0.05 \\

\midrule

\TIES{}
& 75.03 & 84.54 & 87.86 & 94.70 & 85.66 & 82.07 & 98.65 & 67.50 & 84.50 \\
\quad w/ \Surgery{}
& 75.63 $\pm$ 0.17 & 81.36 $\pm$ 0.54 & 88.41 $\pm$ 2.55
& 95.84 $\pm$ 0.54 & 86.21 $\pm$ 0.96 & 83.44 $\pm$ 2.05
& 98.61 $\pm$ 0.17 & 70.46 $\pm$ 2.15 & 84.99 $\pm$ 0.18 \\
\quad w/ \Surgery{V2}
& 78.26 $\pm$ 0.23 & 85.99 $\pm$ 0.51 & 93.92 $\pm$ 1.11
& 99.22 $\pm$ 0.16 & 97.63 $\pm$ 0.07 & 96.52 $\pm$ 0.88
& 99.58 $\pm$ 0.04 & 77.60 $\pm$ 1.04 & 91.09 $\pm$ 0.14 \\
\quad w/ \IntervMerge{}
& 76.85 $\pm$ 0.28 & 83.26 $\pm$ 0.48 & 91.73 $\pm$ 1.87 & 98.27 $\pm$ 0.41 & 96.11 $\pm$ 0.14 & 90.27 $\pm$ 1.52 & 99.36 $\pm$ 0.06 & 74.40 $\pm$ 1.09 & 88.78 $\pm$ 0.19 \\
\quad w/ \FeatCal{}
& \textbf{79.25} $\pm$ 0.06 & \textbf{90.28} $\pm$ 0.07
& 94.02 $\pm$ 0.11 & 99.42 $\pm$ 0.06
& 96.68 $\pm$ 0.03 & 96.59 $\pm$ 0.06
& 99.40 $\pm$ 0.02 & 78.67 $\pm$ 0.20
& 91.79 $\pm$ 0.03 \\
\quad \textbf{w/ \ESCal{}}
& 78.95 $\pm$ 0.16 & 89.25 $\pm$ 0.21
& \textbf{96.01} $\pm$ 0.41 & \textbf{99.59} $\pm$ 0.09
& \textbf{97.90} $\pm$ 0.02 & \textbf{98.55} $\pm$ 0.24
& \textbf{99.61} $\pm$ 0.03 & \textbf{80.39} $\pm$ 0.30
& \textbf{92.53} $\pm$ 0.06 \\

\midrule

\AdaMerging{} (tw)
& 73.63 $\pm$ 2.88 & 81.74 $\pm$ 2.92 & 79.62 $\pm$ 4.97
& 90.39 $\pm$ 2.74 & 87.37 $\pm$ 1.60 & 96.10 $\pm$ 0.31
& 97.48 $\pm$ 0.34 & 70.63 $\pm$ 4.05 & 84.62 $\pm$ 0.66 \\
\quad w/ \Surgery{}
& 74.57 $\pm$ 2.53 & 81.35 $\pm$ 3.45 & 82.71 $\pm$ 4.36
& 94.08 $\pm$ 2.23 & 88.03 $\pm$ 1.42 & 96.21 $\pm$ 0.52
& 97.83 $\pm$ 0.32 & 73.29 $\pm$ 4.13 & 86.01 $\pm$ 0.64 \\
\quad w/ \Surgery{V2}
& 79.77 $\pm$ 1.49 & 88.15 $\pm$ 2.05 & 93.48 $\pm$ 1.80
& 99.44 $\pm$ 0.12 & 97.69 $\pm$ 0.05 & 99.07 $\pm$ 0.10
& 99.54 $\pm$ 0.04 & 81.93 $\pm$ 1.08 & 92.38 $\pm$ 0.28 \\
\quad w/ \IntervMerge{}
& 77.74 $\pm$ 1.31 & 85.16 $\pm$ 2.99 & 88.83 $\pm$ 3.00 & 98.64 $\pm$ 0.52 & 95.75 $\pm$ 0.30 & 98.37 $\pm$ 0.20 & 99.04 $\pm$ 0.11 & 78.41 $\pm$ 3.46 & 90.24 $\pm$ 0.53 \\
\quad w/ \FeatCal{}
& 79.90 $\pm$ 1.60 & \textbf{90.73} $\pm$ 1.15
& 92.30 $\pm$ 1.90 & 99.35 $\pm$ 0.22
& 96.20 $\pm$ 0.26 & 98.80 $\pm$ 0.10
& 99.15 $\pm$ 0.10 & 81.97 $\pm$ 1.22
& 92.30 $\pm$ 0.20 \\
\quad \textbf{w/ \ESCal{}}
& \textbf{79.97} $\pm$ 0.71 & 89.69 $\pm$ 1.07
& \textbf{95.42} $\pm$ 0.57 & \textbf{99.69} $\pm$ 0.04
& \textbf{97.93} $\pm$ 0.06 & \textbf{99.15} $\pm$ 0.05
& \textbf{99.61} $\pm$ 0.03 & \textbf{82.16} $\pm$ 0.93
& \textbf{92.95} $\pm$ 0.11 \\

\midrule

\AdaMerging{} (lw)
& 76.97 $\pm$ 1.29 & 88.59 $\pm$ 1.48 & 88.31 $\pm$ 3.43
& 95.52 $\pm$ 1.18 & 91.38 $\pm$ 2.02 & 97.60 $\pm$ 0.49
& 98.79 $\pm$ 0.40 & 78.39 $\pm$ 1.46 & 89.44 $\pm$ 0.54 \\
\quad w/ \Surgery{}
& 77.79 $\pm$ 1.25 & 88.58 $\pm$ 0.99 & 90.96 $\pm$ 3.21
& 96.81 $\pm$ 0.45 & 91.97 $\pm$ 2.48 & 97.74 $\pm$ 0.35
& 98.82 $\pm$ 0.32 & 79.72 $\pm$ 1.83 & 90.30 $\pm$ 0.60 \\
\quad w/ \Surgery{V2}
& \textbf{81.11} $\pm$ 0.42 & 91.15 $\pm$ 0.41
& 95.90 $\pm$ 0.68 & 99.59 $\pm$ 0.15
& 97.65 $\pm$ 0.10 & 99.13 $\pm$ 0.09
& \textbf{99.61} $\pm$ 0.07 & \textbf{83.40} $\pm$ 0.46
& 93.44 $\pm$ 0.12 \\
\quad w/ \IntervMerge{}
& 80.38 $\pm$ 0.64 & 90.42 $\pm$ 0.78 & 93.61 $\pm$ 2.23 & 98.94 $\pm$ 0.28 & 96.31 $\pm$ 0.35 & 98.46 $\pm$ 0.21 & 99.37 $\pm$ 0.07 & 81.77 $\pm$ 0.61 & 92.41 $\pm$ 0.27 \\
\quad w/ \FeatCal{}
& 80.62 $\pm$ 0.15 & \textbf{91.42} $\pm$ 0.13
& 94.50 $\pm$ 0.71 & 99.39 $\pm$ 0.09
& 95.95 $\pm$ 0.32 & 98.69 $\pm$ 0.09
& 99.32 $\pm$ 0.10 & 83.27 $\pm$ 0.51
& 92.89 $\pm$ 0.11 \\
\quad \textbf{w/ \ESCal{}}
& 80.91 $\pm$ 0.37 & 91.34 $\pm$ 0.29
& \textbf{96.21} $\pm$ 0.30 & \textbf{99.73} $\pm$ 0.08
& \textbf{97.90} $\pm$ 0.03 & \textbf{99.17} $\pm$ 0.03
& \textbf{99.61} $\pm$ 0.04 & 82.83 $\pm$ 0.51
& \textbf{93.46} $\pm$ 0.08 \\

\midrule

\WUDI{}
& 79.65 & 90.95 & 93.98 & 99.41 & 96.21 & 98.04 & 99.63 & 80.64 & 92.31 \\
\quad w/ \Surgery{}
& 79.81 $\pm$ 0.12 & 90.82 $\pm$ 0.12 & 94.65 $\pm$ 0.44
& 99.49 $\pm$ 0.03 & 96.30 $\pm$ 0.12 & 98.11 $\pm$ 0.18
& 99.69 $\pm$ 0.02 & 80.90 $\pm$ 0.60 & 92.47 $\pm$ 0.11 \\
\quad w/ \Surgery{V2}
& \textbf{81.13} $\pm$ 0.12 & 91.36 $\pm$ 0.22
& 96.64 $\pm$ 0.14 & 99.74 $\pm$ 0.04
& 98.06 $\pm$ 0.03 & 99.18 $\pm$ 0.04
& \textbf{99.73} $\pm$ 0.01 & 83.21 $\pm$ 0.37
& 93.63 $\pm$ 0.02 \\
\quad w/ \IntervMerge{}
& 80.62 $\pm$ 0.09 & 91.18 $\pm$ 0.10 & 96.15 $\pm$ 0.47 & 99.65 $\pm$ 0.07 & 97.69 $\pm$ 0.06 & 99.05 $\pm$ 0.09 & 99.71 $\pm$ 0.02 & 82.72 $\pm$ 0.14 & 93.35 $\pm$ 0.06 \\
\quad w/ \FeatCal{}
& 79.92 $\pm$ 0.03 & 90.64 $\pm$ 0.05
& 95.91 $\pm$ 0.07 & 99.74 $\pm$ 0.04
& 97.73 $\pm$ 0.02 & 99.17 $\pm$ 0.02
& 99.70 $\pm$ 0.01 & \textbf{83.91} $\pm$ 0.11
& 93.34 $\pm$ 0.02 \\
\quad \textbf{w/ \ESCal{}}
& 81.08 $\pm$ 0.13 & \textbf{91.56} $\pm$ 0.11
& \textbf{96.76} $\pm$ 0.13 & \textbf{99.79} $\pm$ 0.05
& \textbf{98.14} $\pm$ 0.02 & \textbf{99.21} $\pm$ 0.03
& 99.69 $\pm$ 0.02 & 82.86 $\pm$ 0.21
& \textbf{93.64} $\pm$ 0.04 \\

\midrule

\ProDistill{}
& 77.70 $\pm$ 0.17 & \textbf{89.96} $\pm$ 0.21
& 95.16 $\pm$ 0.67 & 99.56 $\pm$ 0.07
& 97.79 $\pm$ 0.11 & 98.40 $\pm$ 0.34
& 99.60 $\pm$ 0.03 & 78.11 $\pm$ 0.49
& 92.04 $\pm$ 0.07 \\
\quad w/ \Surgery{}
& 77.89 $\pm$ 0.16 & 89.72 $\pm$ 0.14
& 95.18 $\pm$ 0.76 & 99.59 $\pm$ 0.10
& 97.79 $\pm$ 0.10 & 98.39 $\pm$ 0.35
& 99.61 $\pm$ 0.01 & 78.40 $\pm$ 0.18
& 92.07 $\pm$ 0.04 \\
\quad w/ \Surgery{V2}
& 77.98 $\pm$ 0.15 & 89.74 $\pm$ 0.21
& 95.22 $\pm$ 0.69 & 99.56 $\pm$ 0.10
& 97.80 $\pm$ 0.09 & 98.38 $\pm$ 0.36
& 99.62 $\pm$ 0.01 & 78.53 $\pm$ 0.18
& 92.11 $\pm$ 0.06 \\
\quad w/ \IntervMerge{}
& 77.95 $\pm$ 0.15 & 89.78 $\pm$ 0.19 & 95.20 $\pm$ 0.77 & 99.58 $\pm$ 0.09 & 97.80 $\pm$ 0.11 & 98.32 $\pm$ 0.36 & \textbf{99.63} $\pm$ 0.01 & 78.64 $\pm$ 0.36 & 92.11 $\pm$ 0.08 \\
\quad w/ \FeatCal{}
& \textbf{79.14} $\pm$ 0.17 & 89.71 $\pm$ 0.22
& \textbf{96.17} $\pm$ 0.24 & \textbf{99.62} $\pm$ 0.04
& \textbf{97.91} $\pm$ 0.08 & \textbf{98.66} $\pm$ 0.22
& 99.60 $\pm$ 0.03 & \textbf{81.11} $\pm$ 0.42
& \textbf{92.74} $\pm$ 0.10 \\
\quad \textbf{w/ \ESCal{}}
& 77.88 $\pm$ 0.14 & 89.91 $\pm$ 0.26
& 95.32 $\pm$ 0.61 & 99.56 $\pm$ 0.06
& 97.84 $\pm$ 0.09 & 98.46 $\pm$ 0.33
& 99.61 $\pm$ 0.02 & 78.37 $\pm$ 0.18
& 92.12 $\pm$ 0.03 \\

\bottomrule
\end{tabular}
}
\end{table}

\FloatBarrier
\subsection{NLU Results}
\label{app:nlu_results}

Table~\ref{tab:nlu_full} reports the complete task-wise results for
RoBERTa-base across all six base merging methods and the corresponding
post-merging calibration methods.
The \textit{Avg.} column reports the arithmetic mean across the eight NLU tasks,
with the pretrained and expert models included as reference points.

\ESCal{} improves over all uncalibrated models and achieves the highest average score among the compared calibration methods in most merging settings.
The exception is \TIES{}, where \ESCal{} and \Surgery{V2} obtain comparable average scores.

\begin{table}[htbp]
\centering
\renewcommand{\arraystretch}{1.1}
\caption{Full task-wise performance on the NLU benchmark using RoBERTa-base.}
\label{tab:nlu_full}
\resizebox{\linewidth}{!}{
\begin{tabular}{lcccccccc|c}
\toprule
\textbf{Method}
& \textbf{CoLA}
& \textbf{SST-2}
& \textbf{MRPC}
& \textbf{STS-B}
& \textbf{QQP}
& \textbf{MNLI}
& \textbf{QNLI}
& \textbf{RTE}
& \textbf{Avg.} \\
\midrule

Pretrained
& 0.0000 & 0.5975 & 0.1581 & -0.0758 & 0.4633 & 0.3183
& 0.5078 & 0.5271 & 0.3120 \\

Expert
& 0.5895 & 0.9438 & 0.9091 & 0.8997 & 0.8994 & 0.8718
& 0.9198 & 0.7726 & 0.8507 \\

\midrule

\TA{}
& 0.1232 & 0.6135 & 0.5500 & 0.5356 & 0.5881 & 0.6270
& 0.5365 & 0.5343 & 0.5135 \\
\quad w/ \Surgery{}
& 0.2553 $\pm$ 0.0253 & 0.8532 $\pm$ 0.0124
& 0.7657 $\pm$ 0.0155 & 0.7877 $\pm$ 0.0089
& 0.7793 $\pm$ 0.0192 & 0.7215 $\pm$ 0.0118
& 0.7839 $\pm$ 0.0224 & 0.7184 $\pm$ 0.0099
& 0.7081 $\pm$ 0.0088 \\
\quad w/ \Surgery{V2}
& 0.5358 $\pm$ 0.0106 & \textbf{0.9326} $\pm$ 0.0029
& 0.8908 $\pm$ 0.0037 & 0.8978 $\pm$ 0.0018
& 0.8002 $\pm$ 0.0137 & 0.8404 $\pm$ 0.0016
& 0.9084 $\pm$ 0.0019 & 0.7408 $\pm$ 0.0110
& 0.8184 $\pm$ 0.0016 \\
\quad w/ \IntervMerge{}
& 0.4647 $\pm$ 0.0273 & 0.9135 $\pm$ 0.0049 & 0.8477 $\pm$ 0.0061 & 0.8706 $\pm$ 0.0075 & 0.7779 $\pm$ 0.0135 & 0.7625 $\pm$ 0.0080 & 0.8524 $\pm$ 0.0089 & 0.7199 $\pm$ 0.0098 & 0.7762 $\pm$ 0.0067 \\
\quad w/ \FeatCal{}
& 0.3879 $\pm$ 0.0111 & 0.9140 $\pm$ 0.0095
& 0.8162 $\pm$ 0.0116 & 0.4900 $\pm$ 0.0353
& 0.7680 $\pm$ 0.0222 & 0.7569 $\pm$ 0.0026
& 0.8499 $\pm$ 0.0116 & 0.7011 $\pm$ 0.0065
& 0.7105 $\pm$ 0.0024 \\
\quad \textbf{w/ \ESCal{}}
& \textbf{0.5588} $\pm$ 0.0090 & 0.9319 $\pm$ 0.0026
& \textbf{0.8937} $\pm$ 0.0039 & \textbf{0.8998} $\pm$ 0.0019
& \textbf{0.8061} $\pm$ 0.0204 & \textbf{0.8469} $\pm$ 0.0023
& \textbf{0.9117} $\pm$ 0.0029 & \textbf{0.7415} $\pm$ 0.0137
& \textbf{0.8238} $\pm$ 0.0037 \\

\midrule

\TIES{}
& 0.1972 & 0.7626 & 0.5075 & 0.4390 & 0.6039 & 0.6670
& 0.5380 & 0.4946 & 0.5262 \\
\quad w/ \Surgery{}
& 0.2680 $\pm$ 0.0401 & 0.8635 $\pm$ 0.0213
& 0.7570 $\pm$ 0.0082 & 0.7770 $\pm$ 0.0222
& 0.7770 $\pm$ 0.0231 & 0.7347 $\pm$ 0.0130
& 0.7988 $\pm$ 0.0067 & 0.6989 $\pm$ 0.0257
& 0.7094 $\pm$ 0.0126 \\
\quad w/ \Surgery{V2}
& 0.5420 $\pm$ 0.0127 & 0.9319 $\pm$ 0.0021
& \textbf{0.8848} $\pm$ 0.0069 & \textbf{0.8965} $\pm$ 0.0030
& \textbf{0.8205} $\pm$ 0.0117 & \textbf{0.8474} $\pm$ 0.0016
& \textbf{0.9079} $\pm$ 0.0027 & 0.7394 $\pm$ 0.0118
& \textbf{0.8213} $\pm$ 0.0027 \\
\quad w/ \IntervMerge{}
& 0.4382 $\pm$ 0.0255 & 0.9174 $\pm$ 0.0050 & 0.8368 $\pm$ 0.0155 & 0.8702 $\pm$ 0.0096 & 0.8023 $\pm$ 0.0112 & 0.7805 $\pm$ 0.0134 & 0.8660 $\pm$ 0.0049 & 0.7249 $\pm$ 0.0131 & 0.7796 $\pm$ 0.0052 \\
\quad w/ \FeatCal{}
& 0.4055 $\pm$ 0.0130 & 0.9106 $\pm$ 0.0110
& 0.8111 $\pm$ 0.0130 & 0.4765 $\pm$ 0.0498
& 0.7702 $\pm$ 0.0183 & 0.7569 $\pm$ 0.0039
& 0.8433 $\pm$ 0.0115 & 0.7061 $\pm$ 0.0121
& 0.7100 $\pm$ 0.0046 \\
\quad \textbf{w/ \ESCal{}}
& \textbf{0.5552} $\pm$ 0.0253 & \textbf{0.9339} $\pm$ 0.0044
& 0.8769 $\pm$ 0.0113 & 0.8935 $\pm$ 0.0020
& 0.8144 $\pm$ 0.0231 & 0.8443 $\pm$ 0.0024
& 0.9070 $\pm$ 0.0043 & \textbf{0.7430} $\pm$ 0.0103
& 0.8210 $\pm$ 0.0052 \\

\midrule

\AdaMerging{} (tw)
& 0.0283 & 0.4805 & 0.2626 & 0.7031 & 0.2204 & 0.8103
& 0.7818 & 0.7040 & 0.4989 \\
\quad w/ \Surgery{}
& 0.1338 $\pm$ 0.0470 & 0.8234 $\pm$ 0.0293
& 0.7918 $\pm$ 0.0095 & 0.7757 $\pm$ 0.0188
& 0.7438 $\pm$ 0.0072 & 0.8336 $\pm$ 0.0034
& 0.8711 $\pm$ 0.0066 & 0.7401 $\pm$ 0.0068
& 0.7142 $\pm$ 0.0076 \\
\quad w/ \Surgery{V2}
& 0.5227 $\pm$ 0.0259 & 0.9339 $\pm$ 0.0056
& 0.8850 $\pm$ 0.0079 & \textbf{0.8910} $\pm$ 0.0031
& 0.7570 $\pm$ 0.0195 & \textbf{0.8651} $\pm$ 0.0015
& \textbf{0.9151} $\pm$ 0.0015 & \textbf{0.7783} $\pm$ 0.0253
& 0.8185 $\pm$ 0.0071 \\
\quad w/ \IntervMerge{}
& 0.2981 $\pm$ 0.0591 & 0.8945 $\pm$ 0.0286 & 0.8013 $\pm$ 0.0285 & 0.8272 $\pm$ 0.0368 & 0.7287 $\pm$ 0.0193 & 0.8507 $\pm$ 0.0058 & 0.8763 $\pm$ 0.0171 & 0.7422 $\pm$ 0.0146 & 0.7524 $\pm$ 0.0096 \\
\quad w/ \FeatCal{}
& 0.4004 $\pm$ 0.0193 & 0.9183 $\pm$ 0.0075
& 0.8189 $\pm$ 0.0057 & 0.4628 $\pm$ 0.0479
& \textbf{0.7737} $\pm$ 0.0164 & 0.7525 $\pm$ 0.0032
& 0.8194 $\pm$ 0.0106 & 0.6606 $\pm$ 0.0142
& 0.7008 $\pm$ 0.0042 \\
\quad \textbf{w/ \ESCal{}}
& \textbf{0.5383} $\pm$ 0.0163 & \textbf{0.9344} $\pm$ 0.0036
& \textbf{0.8892} $\pm$ 0.0074 & 0.8905 $\pm$ 0.0023
& 0.7594 $\pm$ 0.0243 & 0.8648 $\pm$ 0.0017
& 0.9110 $\pm$ 0.0028 & 0.7740 $\pm$ 0.0083
& \textbf{0.8202} $\pm$ 0.0048 \\

\midrule

\AdaMerging{} (lw)
& 0.0137 & 0.5115 & 0.3069 & 0.7205 & 0.1868 & 0.8038
& 0.6077 & 0.7076 & 0.4823 \\
\quad w/ \Surgery{}
& 0.1864 $\pm$ 0.0594 & 0.8314 $\pm$ 0.0149
& 0.7948 $\pm$ 0.0099 & 0.7699 $\pm$ 0.0338
& 0.7763 $\pm$ 0.0111 & 0.8182 $\pm$ 0.0077
& 0.8242 $\pm$ 0.0106 & 0.7372 $\pm$ 0.0181
& 0.7173 $\pm$ 0.0122 \\
\quad w/ \Surgery{V2}
& 0.5145 $\pm$ 0.0201 & 0.9333 $\pm$ 0.0074
& 0.8867 $\pm$ 0.0055 & 0.8930 $\pm$ 0.0031
& 0.8073 $\pm$ 0.0204 & 0.8585 $\pm$ 0.0008
& 0.9080 $\pm$ 0.0028 & \textbf{0.7711} $\pm$ 0.0197
& 0.8215 $\pm$ 0.0059 \\
\quad w/ \IntervMerge{}
& 0.4303 $\pm$ 0.0270 & 0.9147 $\pm$ 0.0100 & 0.8324 $\pm$ 0.0280 & 0.8544 $\pm$ 0.0134 & 0.7620 $\pm$ 0.0146 & 0.8355 $\pm$ 0.0119 & 0.8510 $\pm$ 0.0196 & 0.7357 $\pm$ 0.0086 & 0.7770 $\pm$ 0.0052 \\
\quad w/ \FeatCal{}
& 0.4052 $\pm$ 0.0288 & 0.9092 $\pm$ 0.0136
& 0.8297 $\pm$ 0.0133 & 0.4788 $\pm$ 0.0306
& 0.7767 $\pm$ 0.0181 & 0.7368 $\pm$ 0.0027
& 0.8354 $\pm$ 0.0092 & 0.6801 $\pm$ 0.0261
& 0.7065 $\pm$ 0.0030 \\
\quad \textbf{w/ \ESCal{}}
& \textbf{0.5391} $\pm$ 0.0140 & \textbf{0.9339} $\pm$ 0.0029
& \textbf{0.8926} $\pm$ 0.0057 & \textbf{0.8968} $\pm$ 0.0010
& \textbf{0.8204} $\pm$ 0.0175 & \textbf{0.8592} $\pm$ 0.0015
& \textbf{0.9082} $\pm$ 0.0035 & 0.7704 $\pm$ 0.0119
& \textbf{0.8276} $\pm$ 0.0040 \\

\midrule

\WUDI{}
& 0.0625 & 0.8073 & 0.2565 & 0.0927 & 0.7680 & 0.4431
& 0.4622 & 0.4874 & 0.4225 \\
\quad w/ \Surgery{}
& 0.2226 $\pm$ 0.0296 & 0.9025 $\pm$ 0.0095
& 0.7860 $\pm$ 0.0111 & 0.7758 $\pm$ 0.0264
& 0.8270 $\pm$ 0.0054 & 0.7755 $\pm$ 0.0179
& 0.8118 $\pm$ 0.0165 & 0.7206 $\pm$ 0.0139
& 0.7277 $\pm$ 0.0059 \\
\quad w/ \Surgery{V2}
& 0.5529 $\pm$ 0.0130 & 0.9392 $\pm$ 0.0049
& 0.8908 $\pm$ 0.0087 & 0.8868 $\pm$ 0.0027
& \textbf{0.8798} $\pm$ 0.0032 & \textbf{0.8620} $\pm$ 0.0017
& 0.9136 $\pm$ 0.0016 & \textbf{0.7603} $\pm$ 0.0124
& 0.8357 $\pm$ 0.0018 \\
\quad w/ \IntervMerge{}
& 0.4766 $\pm$ 0.0087 & 0.9282 $\pm$ 0.0050 & 0.8303 $\pm$ 0.0135 & 0.8668 $\pm$ 0.0102 & 0.8562 $\pm$ 0.0027 & 0.8347 $\pm$ 0.0054 & 0.8884 $\pm$ 0.0021 & 0.7516 $\pm$ 0.0112 & 0.8041 $\pm$ 0.0027 \\
\quad w/ \FeatCal{}
& 0.4387 $\pm$ 0.0245 & 0.9190 $\pm$ 0.0084
& 0.8215 $\pm$ 0.0082 & 0.4953 $\pm$ 0.0427
& 0.8234 $\pm$ 0.0112 & 0.7748 $\pm$ 0.0033
& 0.8358 $\pm$ 0.0132 & 0.6888 $\pm$ 0.0213
& 0.7247 $\pm$ 0.0069 \\
\quad \textbf{w/ \ESCal{}}
& \textbf{0.5663} $\pm$ 0.0217 & \textbf{0.9397} $\pm$ 0.0055
& \textbf{0.8950} $\pm$ 0.0051 & \textbf{0.8953} $\pm$ 0.0027
& 0.8744 $\pm$ 0.0061 & 0.8616 $\pm$ 0.0013
& \textbf{0.9144} $\pm$ 0.0025 & 0.7545 $\pm$ 0.0120
& \textbf{0.8376} $\pm$ 0.0024 \\

\midrule

\ProDistill{} (ew)
& 0.4076 & 0.9106 & 0.8463 & 0.5288 & 0.7938 & 0.7759
& 0.8609 & 0.7148 & 0.7298 \\
\quad w/ \Surgery{}
& 0.4470 $\pm$ 0.0225 & 0.9158 $\pm$ 0.0071
& 0.8371 $\pm$ 0.0075 & 0.7119 $\pm$ 0.0494
& 0.7756 $\pm$ 0.0253 & 0.7851 $\pm$ 0.0061
& 0.8647 $\pm$ 0.0032 & 0.7249 $\pm$ 0.0164
& 0.7578 $\pm$ 0.0049 \\
\quad w/ \Surgery{V2}
& \textbf{0.5006} $\pm$ 0.0219 & \textbf{0.9257} $\pm$ 0.0042
& 0.8717 $\pm$ 0.0074 & 0.8001 $\pm$ 0.0274
& \textbf{0.8089} $\pm$ 0.0233 & 0.8283 $\pm$ 0.0032
& 0.9012 $\pm$ 0.0023 & \textbf{0.7256} $\pm$ 0.0137
& 0.7953 $\pm$ 0.0051 \\
\quad w/ \IntervMerge{}
& 0.4664 $\pm$ 0.0119 & 0.9202 $\pm$ 0.0084 & 0.8582 $\pm$ 0.0056 & 0.7711 $\pm$ 0.0188 & 0.7986 $\pm$ 0.0234 & 0.7965 $\pm$ 0.0065 & 0.8840 $\pm$ 0.0048 & \textbf{0.7256} $\pm$ 0.0135 & 0.7776 $\pm$ 0.0030 \\
\quad w/ \FeatCal{}
& 0.4384 $\pm$ 0.0244 & 0.9122 $\pm$ 0.0045
& 0.8246 $\pm$ 0.0148 & 0.4523 $\pm$ 0.0716
& 0.8062 $\pm$ 0.0160 & 0.7914 $\pm$ 0.0074
& 0.8493 $\pm$ 0.0138 & 0.7141 $\pm$ 0.0160
& 0.7235 $\pm$ 0.0116 \\
\quad \textbf{w/ \ESCal{}}
& 0.4880 $\pm$ 0.0297 & 0.9239 $\pm$ 0.0045
& \textbf{0.8759} $\pm$ 0.0069 & \textbf{0.8734} $\pm$ 0.0106
& 0.7913 $\pm$ 0.0388 & \textbf{0.8292} $\pm$ 0.0023
& \textbf{0.9017} $\pm$ 0.0027 & 0.7155 $\pm$ 0.0179
& \textbf{0.7999} $\pm$ 0.0062 \\

\bottomrule
\end{tabular}
}
\end{table}

\subsection{NLG Results}
\label{app:nlg_results}
We further evaluate post-merging calibration on natural language generation (NLG) using LLaMA-2-13B experts specialized for instruction following, mathematical reasoning, and code generation, considering all three pairwise merges.

\paragraph{Aggregation and expert references.}
For each expert pair, Avg. is the arithmetic mean over all benchmarks
shown in the corresponding table, including AlpacaEval when present.
Norm Avg. divides each benchmark score by the score of the corresponding
calibration target expert and then averages these ratios.
For benchmark set $\mathcal B$, we report
\begin{equation}
    \mathrm{Avg.}(m)=\frac{1}{|\mathcal B|}\sum_{b\in\mathcal B}s_{m,b},
    \qquad
    \mathrm{Norm\ Avg.}(m)=\frac{1}{|\mathcal B|}
    \sum_{b\in\mathcal B}\frac{s_{m,b}}{s_{e(b),b}},
\end{equation}
where $s_{m,b}$ is the score of model $m$ on benchmark $b$, and $e(b)$
is the designated calibration target expert for that benchmark.
The reference is fixed across all rows within each table.
A per-benchmark normalized score above one indicates performance above
the corresponding target expert; a normalized average above one need
not indicate improvements on every benchmark.

For math--code merging, WizardMath-13B is the reference for GSM8K and
MATH, while Llama-2-13b-code-alpaca is the reference for HumanEval and MBPP.
For instruction--code merging, Llama-2-13b-code-alpaca is the reference
for HumanEval and MBPP, and WizardLM-13B-V1.2 is the reference for AlpacaEval.
For instruction--math merging, WizardMath-13B is the reference for GSM8K
and MATH, while WizardLM-13B-V1.2 is the reference for HumanEval, MBPP,
and AlpacaEval.
The reference therefore follows the calibration target rather than
whichever constituent expert obtains the higher benchmark score.
The reported aggregates are computed from the displayed benchmark scores
and rounded to four decimal places.

\begin{table*}[htbp]
\centering
\caption{Post-merging calibration for LLaMA-2-13B math--code merging.}
\label{tab:nlg_math_code}
\resizebox{0.8\textwidth}{!}{
\begin{tabular}{lcccc|cc}
\toprule
\textbf{Method}
& \textbf{GSM8K}
& \textbf{MATH}
& \textbf{HumanEval}
& \textbf{MBPP}
& \textbf{Avg.}
& \textbf{Norm Avg.} \\
\midrule

WizardMath-13B
& 0.6343 & 0.1468 & 0.0732 & 0.2156 & 0.2675 & 0.7655 \\
Llama-2-13b-code-alpaca
& 0.0000 & 0.0000 & 0.2195 & 0.2959 & 0.1289 & 0.5000 \\

\midrule

Task Arithmetic
& 0.6494 & 0.1446 & 0.0122 & 0.0917 & 0.2245 & 0.5936 \\
\quad w/ Surgery
& 0.6542 & 0.1468 & 0.0488 & 0.1514 & 0.2503 & 0.6913 \\
\quad w/ \Surgery{V2}
& 0.6255 & 0.1430 & 0.0793 & 0.2133 & 0.2653 & 0.7606 \\
\quad w/ \FeatCal{}
& 0.6406 & 0.1452 & 0.2073 & 0.2683 & \textbf{0.3154} & \textbf{0.9625} \\
\quad \textbf{w/ \method{ESCal}}
& 0.6351 & 0.1544 & 0.1829 & 0.2523 & 0.3062 & 0.9347 \\

\midrule

TIES-Merging
& 0.5912 & 0.1194 & 0.0976 & 0.1330 & 0.2353 & 0.6599 \\
\quad w/ Surgery
& 0.6159 & 0.1202 & 0.0976 & 0.1628 & 0.2491 & 0.6962 \\
\quad w/ \Surgery{V2}
& 0.5976 & 0.1180 & 0.1890 & 0.2867 & 0.2978 & 0.8940 \\
\quad w/ \FeatCal{}
& 0.6263 & 0.1352 & 0.1829 & 0.2569 & 0.3003 & 0.9025 \\
\quad \textbf{w/ \method{ESCal}}
& 0.6112 & 0.1298 & 0.1951 & 0.2798 & \textbf{0.3040} & \textbf{0.9206} \\

\bottomrule
\end{tabular}
}
\end{table*}

\begin{table*}[htbp]
\centering
\caption{Post-merging calibration for LLaMA-2-13B instruction--code merging.}
\label{tab:nlg_ins_code}
\resizebox{0.8\textwidth}{!}{
\begin{tabular}{lccc|cc}
\toprule
\textbf{Method}
& \textbf{HumanEval}
& \textbf{MBPP}
& \textbf{AlpacaEval}
& \textbf{Avg.}
& \textbf{Norm Avg.} \\
\midrule

WizardLM-13B-V1.2
& 0.3293 & 0.3509 & 0.1646 & 0.2816 & 1.2287 \\
Llama-2-13b-code-alpaca
& 0.2195 & 0.2959 & 0.0196 & 0.1783 & 0.7064 \\

\midrule

Task Arithmetic
& 0.3232 & 0.3440 & 0.1576 & 0.2749 & 1.1975 \\
\quad w/ Surgery
& 0.3171 & 0.3486 & 0.1604 & \textbf{0.2754} & 1.1991 \\
\quad w/ \Surgery{V2}
& 0.3293 & 0.3349 & 0.1615 & 0.2752 & \textbf{1.2044} \\
\quad w/ \FeatCal{}
& 0.3110 & 0.3372 & 0.1520 & 0.2667 & 1.1600 \\
\quad \textbf{w/ \method{ESCal}}
& 0.3110 & 0.3417 & 0.1570 & 0.2699 & 1.1752 \\

\midrule

TIES-Merging
& 0.2317 & 0.3326 & 0.0501 & 0.2048 & 0.8280 \\
\quad w/ Surgery
& 0.2561 & 0.3394 & 0.0527 & 0.2161 & 0.8780 \\
\quad w/ \Surgery{V2}
& 0.2622 & 0.3372 & 0.0601 & 0.2198 & 0.8997 \\
\quad w/ \FeatCal{}
& 0.2744 & 0.3211 & 0.0622 & 0.2192 & 0.9044 \\
\quad \textbf{w/ \method{ESCal}}
& 0.2805 & 0.3326 & 0.0819 & \textbf{0.2317} & \textbf{0.9665} \\

\bottomrule
\end{tabular}
}
\end{table*}

\begin{table*}[htbp]
\centering
\caption{Post-merging calibration for LLaMA-2-13B instruction--math merging.}
\label{tab:nlg_ins_math}
\resizebox{0.8\textwidth}{!}{
\begin{tabular}{lccccc|cc}
\toprule
\textbf{Method}
& \textbf{GSM8K}
& \textbf{MATH}
& \textbf{HumanEval}
& \textbf{MBPP}
& \textbf{AlpacaEval}
& \textbf{Avg.}
& \textbf{Norm Avg.} \\
\midrule

WizardLM-13B-V1.2
& 0.0223 & 0.0004 & 0.3293 & 0.3280 & 0.1646 & 0.1689 & 0.6076 \\
WizardMath-13B
& 0.6343 & 0.1468 & 0.0732 & 0.2100 & 0.0487 & 0.2226 & 0.6317 \\

\midrule

Task Arithmetic
& 0.6614 & 0.1440 & 0.2622 & 0.3000 & 0.1053 & 0.2946 & 0.8749 \\
\quad w/ Surgery
& 0.6693 & 0.1358 & 0.2683 & 0.3300 & 0.1101 & \textbf{0.3027} & \textbf{0.8940} \\
\quad w/ \Surgery{V2}
& 0.6566 & 0.1434 & 0.2500 & 0.3160 & 0.1149 & 0.2962 & 0.8865 \\
\quad w/ \FeatCal{}
& 0.6454 & 0.1404 & 0.2439 & 0.3040 & 0.1041 & 0.2876 & 0.8548 \\
\quad \textbf{w/ \method{ESCal}}
& 0.6653 & 0.1450 & 0.2561 & 0.2980 & 0.1073 & 0.2943 & 0.8749 \\

\midrule

TIES-Merging
& 0.5761 & 0.1042 & 0.0915 & 0.3080 & 0.0609 & 0.2281 & 0.6410 \\
\quad w/ Surgery
& 0.6008 & 0.1142 & 0.1951 & 0.3240 & 0.0899 & 0.2648 & 0.7703 \\
\quad w/ \Surgery{V2}
& 0.6064 & 0.1156 & 0.2317 & 0.3220 & 0.0920 & 0.2735 & 0.7975 \\
\quad w/ \FeatCal{}
& 0.6375 & 0.1366 & 0.1890 & 0.3020 & 0.1095 & 0.2749 & \textbf{0.8191} \\
\quad \textbf{w/ \method{ESCal}}
& 0.6104 & 0.1252 & 0.2378 & 0.3340 & 0.0859 & \textbf{0.2787} & 0.8155 \\

\bottomrule
\end{tabular}
}
\end{table*}
\paragraph{Calibration when the merged model already performs well.}
Task Arithmetic already outperforms its calibration targets on some
benchmarks.
For example, Task Arithmetic achieves a GSM8K score of 0.6614 in the
instruction--math merge, exceeding WizardMath-13B's 0.6343.
In instruction--code merging, its HumanEval and MBPP scores also exceed
those of the code expert, although they remain below those of WizardLM-13B-V1.2.
Its normalized average is 1.1975 relative to the designated targets,
although its AlpacaEval score remains below that of the instruction expert.
A merged model can also exceed both experts' overall averages by combining
their strengths, even if it does not outperform the stronger expert on every benchmark.

\Surgery{}, \Surgery{V2}, \FeatCal{}, and \ESCal{} aim to reduce
merging-induced discrepancies by using the constituent experts as
calibration targets or references.
When merging already improves a benchmark score over the corresponding
expert, closer alignment with that expert need not yield further
performance gains.
Calibration may still improve other benchmarks. For Task Arithmetic,
its effect on the overall average varies across calibration methods.
In instruction--code merging, the full-benchmark average changes from
0.2749 to 0.2754 with \Surgery{} and 0.2752 with \Surgery{V2}, but to
0.2667 with \FeatCal{} and 0.2699 with \ESCal{}.

\paragraph{Calibration when merging leaves an expert-performance gap.}
The TIES merges remain below their designated calibration targets on
average across all three expert pairs.
\ESCal{} improves both Avg. and Norm Avg. for all three TIES merges,
and achieves the highest values for both metrics among the evaluated
calibration variants in the math--code and instruction--code settings.
For math--code merging, it increases Avg. from 0.2353 to 0.3040 and
Norm Avg. from 0.6599 to 0.9206.
For instruction--code merging, the corresponding improvements are
0.2048 to 0.2317 and 0.8280 to 0.9665.
These results show that corrections between fixed error subspaces can
improve performance when the merged model remains below its calibration
targets.

\FloatBarrier
\section{Additional Analyses}
\label{app:additional_analyses}

\subsection{Error-Subspace Construction}
\label{app:error_subspace_details}

Table~\ref{tab:vision_ablation_basis_full} provides the full per-task
results for the basis-source comparison in
Table~\ref{tab:vision_ablation_basis}, together with the bottom-$r$
merging-error baseline.
Bases derived from the top-$r$ singular directions of the merging error
achieve the highest mean accuracy on all eight tasks.
Their average accuracy is $87.54\%$, compared with $87.01\%$ for task-vector
bases and $84.80\%$ for fine-tuned-weight bases.
Random orthonormal bases and bottom-$r$ merging-error bases yield
$84.08\%$ and $80.86\%$, respectively.
These results support using the dominant singular subspaces of the
expert--merged weight discrepancy to constrain calibration.

\begin{table}[htbp]
\centering
\renewcommand{\arraystretch}{1.25}
\caption{
\textbf{Full results for the error-subspace construction ablation.}
We report per-task and average accuracy (\%; mean $\pm$ standard deviation)
for the basis-source comparison in
Table~\ref{tab:vision_ablation_basis}, together with the bottom-$r$
merging-error baseline.
}

\label{tab:vision_ablation_basis_full}
\renewcommand{\arraystretch}{1.25}
\resizebox{\textwidth}{!}{
\begin{tabular}{lcccccccc|c}
\toprule
\textbf{Basis construction}
& \textbf{SUN397}
& \textbf{Cars}
& \textbf{RESISC45}
& \textbf{EuroSAT}
& \textbf{SVHN}
& \textbf{GTSRB}
& \textbf{MNIST}
& \textbf{DTD}
& \textbf{Avg.} \\
\midrule
Random orthonormal
& 65.42 $\pm$ 0.27 & 63.88 $\pm$ 0.55 & 88.54 $\pm$ 0.57
& 98.61 $\pm$ 0.23 & 96.06 $\pm$ 0.10 & 95.81 $\pm$ 0.22
& 99.32 $\pm$ 0.05 & 64.98 $\pm$ 1.02 & 84.08 $\pm$ 0.15 \\
\midrule
Fine-tuned weights, top-$r$
& 66.94 $\pm$ 0.22 & 65.02 $\pm$ 0.48 & 89.85 $\pm$ 0.61
& 99.03 $\pm$ 0.15 & 96.17 $\pm$ 0.04 & 96.32 $\pm$ 0.45
& 99.36 $\pm$ 0.06 & 65.70 $\pm$ 0.84 & 84.80 $\pm$ 0.24 \\
Task vector, top-$r$
& 69.29 $\pm$ 0.15 & 67.72 $\pm$ 0.67 & 93.01 $\pm$ 0.45
& 99.44 $\pm$ 0.11 & 97.08 $\pm$ 0.04 & 97.47 $\pm$ 0.28
& 99.53 $\pm$ 0.04 & 72.53 $\pm$ 0.39 & 87.01 $\pm$ 0.09 \\
\midrule
Merging error, bottom-$r$
& 62.11 $\pm$ 0.21 & 62.17 $\pm$ 0.30 & 82.26 $\pm$ 0.36
& 97.20 $\pm$ 0.35 & 94.02 $\pm$ 0.19 & 91.05 $\pm$ 0.17
& 98.83 $\pm$ 0.07 & 59.21 $\pm$ 0.58 & 80.86 $\pm$ 0.13 \\
\textbf{Merging error, top-$r$}
& \textbf{70.33} $\pm$ 0.22 & \textbf{69.02} $\pm$ 0.19 & \textbf{93.60} $\pm$ 0.36
& \textbf{99.48} $\pm$ 0.07 & \textbf{97.10} $\pm$ 0.05 & \textbf{97.70} $\pm$ 0.22
& \textbf{99.55} $\pm$ 0.03 & \textbf{73.52} $\pm$ 0.44 & \textbf{87.54} $\pm$ 0.07 \\
\bottomrule
\end{tabular}
}
\end{table}

\FloatBarrier
\subsection{Computational Cost}
\label{app:computational_cost}

We evaluate trainable parameter counts, additional inference-time storage,
and computation time on ViT-B/32 with $T=8$ tasks and 64 calibration
examples per task. Storage and inference measurements assume a single
active task.

\paragraph{Parameters and storage.}
Table~\ref{tab:parameter_storage_cost} summarizes the trainable parameters
across all eight tasks and the additional tensor storage required for
single-task inference, relative to the shared image encoder.
We use $L=12$, $d=768$, $d_{\mathrm{out}}=512$, $r_s=16$, and $r_b=128$.

\begin{table}[htbp]
\centering
\caption{
Trainable parameters across eight tasks and additional storage for
single-task inference on ViT-B/32. 
}

\label{tab:parameter_storage_cost}
\renewcommand{\arraystretch}{1.1}
\resizebox{\linewidth}{!}{
\begin{tabular}{lcccccc}
\toprule
& \multicolumn{2}{c}{\textbf{Training ($T=8$ tasks)}}
& \multicolumn{4}{c}{\textbf{Inference (one task)}} \\
\cmidrule(lr){2-3}\cmidrule(lr){4-7}
\textbf{Method}
& \textbf{Trainable Parameters}
& \textbf{\# Trainable}
& \textbf{Stored Entries}
& \textbf{\# Stored}
& \textbf{BF16 Storage}
& \textbf{Overhead} \\
\midrule
Shared ViT-B/32 image encoder
& -- & -- & $P_{\mathrm{img}}$ & 87,849,216 & 167.56 MiB & Reference \\
\midrule
\Surgery{}
& $2T d_{\mathrm{out}}r_s$ & 131,072
& $2d_{\mathrm{out}}r_s$ & 16,384
& 0.031 MiB & 0.02\% \\
\Surgery{V2}
& $T(2Ldr_s+2d_{\mathrm{out}}r_s)$ & 2,490,368
& $2Ldr_s+2d_{\mathrm{out}}r_s$ & 311,296
& 0.594 MiB & 0.35\% \\
\FeatCal{}
& $0$ (closed form) & 0
& $0$ & 0
& 0.000 MiB & 0.00\% \\
\textbf{\ESCal{}}
& $TLr_b^2$ & 1,572,864
& $Lr_b^2+2Ldr_b$ & 2,555,904
& 4.875 MiB & 2.91\% \\
\bottomrule
\end{tabular}
}
\end{table}

\ESCal{} learns $TLr_b^2=1{,}572{,}864$ core parameters across eight tasks,
which is $36.8\%$ fewer than \Surgery{V2}.
For single-task inference, it stores $Lr_b^2=196{,}608$ core entries and
$2Ldr_b=2{,}359{,}296$ fixed read/write basis entries.
The resulting 2,555,904 additional entries require 4.875 MiB at BF16,
equivalent to $2.91\%$ of the shared image encoder's storage.
Thus, \ESCal{} reduces the number of trainable parameters relative to
\Surgery{V2}, while requiring more inference-time storage for its fixed bases.

\Surgery{} applies low-rank calibration at the output embedding, while
\Surgery{V2} additionally applies calibration at each residual block.
\FeatCal{} updates the merged encoder in closed form and stores the
resulting weights in place, requiring no additional inference-time tensors
for the active task.
This accounting excludes activations, optimizer states, temporary buffers,
and serialization metadata; it does not quantify peak training memory
or total training computation.

\paragraph{Construction and calibration time.}
Table~\ref{tab:wallclock_cost} reports base-model construction and
calibration times separately.
We compare the time needed for base merging followed by calibration with
that of more expensive merging procedures.
The total time is the sum of base-model construction and calibration.
Stage timings include setup, training or closed-form updates, and saving
model artifacts, but exclude test evaluation, library imports, and reporting.

\begin{table}[htbp]
\centering
\caption{
Base-model construction and calibration times in minutes on ViT-B/32 with eight tasks and 64 calibration examples per task (single run on an NVIDIA RTX A6000).
Calibration times exclude base-model construction; gradient-based calibration uses 200 epochs.
The task-wise and layer-wise variants of \AdaMerging{} are denoted by
\textnormal{tw} and \textnormal{lw}, respectively.
}

\label{tab:wallclock_cost}
\resizebox{0.9\linewidth}{!}{
\begin{tabular}{lccccc}
\toprule
& \textbf{Base construction}
& \multicolumn{4}{c}{\textbf{Calibration}} \\
\cmidrule(lr){2-2}\cmidrule(lr){3-6}
\textbf{Base merging method}
& \textbf{Time}
& \textbf{\Surgery{}}
& \textbf{\Surgery{V2}}
& \textbf{\FeatCal{}}
& \textbf{\ESCal{}} \\
\midrule
\TA{}
& 0.09 & 6.28 & 13.41 & 0.63 & 13.22 \\
\TIES{}
& 0.39 & 5.48 & 12.05 & 0.58 & 12.92 \\
\WUDI{}
& 5.61 & 5.53 & 12.18 & 0.53 & 13.05 \\
\AdaMerging{} (tw)
& 39.46 & 6.87 & 13.69 & 0.65 & 14.06 \\
\AdaMerging{} (lw)
& 40.04 & 6.96 & 13.69 & 0.89 & 14.05 \\
\ProDistill{}
& 23.26 & 6.73 & 13.38 & 0.61 & 14.05 \\
\bottomrule
\end{tabular}
}
\end{table}

For this timing comparison, we use 200 calibration epochs for all
gradient-based methods: 400 outer iterations for \Surgery{} and \Surgery{V2},
and 200 epochs for \ESCal{}.
With 64 calibration examples per task and a batch size of 16,
each method performs 800 updates per task.
This timing protocol is separate from the best-setting accuracy evaluation.
\FeatCal{} uses the same calibration examples for its closed-form update.
Base-model construction uses 500 epochs for \AdaMerging{} and 100 epochs
per block for \ProDistill{}; for \TA{}, it includes loading, merging,
and saving the model.

\paragraph{Inference latency and throughput.}
We compare inference cost with and without calibration modules on a
\TA{} ViT-B/32 model.
As shown in Table~\ref{tab:inference_cost_single_checkpoint},
\ESCal{} increases per-image latency by $28.53\%$ at batch size 1
and reduces throughput by $5.29\%$ at batch size 64.
These overheads are similar to those of \Surgery{V2}
($31.91\%$ and $5.04\%$, respectively) in this setting.

We benchmark a model merged with coefficient 0.3, using SUN397 as
the active task, on an NVIDIA RTX A6000 with BF16 autocast.
Timing covers the image encoder, calibration modules, and classification
head, with preprocessed inputs already on the GPU.
We perform 10 paired repetitions of each calibrated model and its
uncalibrated counterpart, alternating their execution order.
Each repetition uses 100 warm-up passes and 200 timed passes
with CUDA synchronization.

\begin{table}[htbp]
\centering
\caption{
\textbf{Inference cost relative to the uncalibrated \TA{} ViT-B/32 model.}
\textnormal{None} denotes the model without calibration.
Absolute values are mean $\pm$ standard deviation.
$\Delta t$ denotes additional per-image latency.
Relative changes are computed against paired uncalibrated runs
and reported with 95\% bootstrap confidence intervals;
they may therefore differ from ratios based on the pooled
\textnormal{None} row.
}
\label{tab:inference_cost_single_checkpoint}
\resizebox{\linewidth}{!}{
\begin{tabular}{lcccccc}
\toprule
& \multicolumn{3}{c}{\textbf{Batch size 1}}
& \multicolumn{3}{c}{\textbf{Batch size 64}} \\
\cmidrule(lr){2-4}\cmidrule(lr){5-7}
\textbf{Calibration} & \textbf{Image latency (ms)} & $\Delta t$ \textbf{(ms)}
& \textbf{Overhead (\%)} & \textbf{Batch latency (ms)}
& \textbf{Throughput (img/s)} & \textbf{Throughput reduction (\%)} \\
\midrule
None & 5.51 $\pm$ 0.02 & Ref. & Ref. & 14.83 $\pm$ 0.05 & 4316.5 $\pm$ 15.6 & Ref. \\
\Surgery{} & 5.60 $\pm$ 0.03 & +0.095 & +1.72 [+1.51, +1.95] & 14.81 $\pm$ 0.07 & 4320.1 $\pm$ 20.0 & +0.15 [+0.06, +0.25] \\
\Surgery{V2} & 7.27 $\pm$ 0.02 & +1.758 & +31.91 [+31.60, +32.27] & 15.63 $\pm$ 0.04 & 4095.0 $\pm$ 11.6 & +5.04 [+4.99, +5.08] \\
\FeatCal{} & 5.50 $\pm$ 0.02 & +0.004 & +0.07 [-0.30, +0.48] & 14.84 $\pm$ 0.03 & 4311.4 $\pm$ 8.3 & +0.01 [-0.01, +0.04] \\
\textbf{\ESCal{}} & 7.09 $\pm$ 0.02 & +1.575 & +28.53 [+28.23, +28.84] & 15.66 $\pm$ 0.04 & 4086.9 $\pm$ 10.3 & +5.29 [+5.22, +5.35] \\
\bottomrule
\end{tabular}
}
\end{table}

\subsection{Calibration Data Budget}
\label{app:sample_efficiency}
We evaluate how \ESCal{} performance changes with the number of
calibration examples per task on ViT-B/32 with \TA{} as the base
merging method, using 64 examples as the reference.
For each data budget, we use the same merged checkpoint and vary only
the task-specific calibration examples.
We report task-wise accuracy and its average across tasks.

\begin{table}[htbp]
\centering
\renewcommand{\arraystretch}{1.1}
\caption{Effect of the per-task support size on \ESCal{}, ViT-B/32 with a task-arithmetic base. Epochs are held at 200, so the optimizer's update count scales with the support size. Accuracy is on the full test set; the default setting ($64$) is set in bold.}
\label{tab:vision_ablation_valshot}
\resizebox{\textwidth}{!}{
\begin{tabular}{lcccccccc|c}
\toprule
\textbf{Support / task}
& \textbf{SUN397}
& \textbf{Cars}
& \textbf{RESISC45}
& \textbf{EuroSAT}
& \textbf{SVHN}
& \textbf{GTSRB}
& \textbf{MNIST}
& \textbf{DTD}
& \textbf{Avg.} \\
\midrule
4
& 66.64 $\pm$ 0.21 & 63.17 $\pm$ 1.20 & 85.49 $\pm$ 1.08
& 91.32 $\pm$ 3.07 & 93.87 $\pm$ 0.33 & 92.01 $\pm$ 1.37
& 97.47 $\pm$ 0.49 & 60.33 $\pm$ 1.44 & 81.29 $\pm$ 0.36 \\
8
& 67.83 $\pm$ 0.42 & 65.05 $\pm$ 1.21 & 88.01 $\pm$ 0.91
& 94.77 $\pm$ 3.28 & 95.47 $\pm$ 0.33 & 93.60 $\pm$ 0.75
& 98.27 $\pm$ 0.64 & 63.48 $\pm$ 1.61 & 83.31 $\pm$ 0.49 \\
16
& 68.71 $\pm$ 0.27 & 66.50 $\pm$ 0.84 & 90.03 $\pm$ 0.56
& 96.72 $\pm$ 0.91 & 96.02 $\pm$ 0.51 & 94.87 $\pm$ 0.69
& 98.79 $\pm$ 0.34 & 65.49 $\pm$ 0.86 & 84.64 $\pm$ 0.34 \\
32
& 69.68 $\pm$ 0.14 & 67.77 $\pm$ 0.41 & 92.35 $\pm$ 0.28
& 99.09 $\pm$ 0.23 & 96.79 $\pm$ 0.11 & 96.90 $\pm$ 0.63
& 99.38 $\pm$ 0.07 & 69.89 $\pm$ 1.42 & 86.48 $\pm$ 0.21 \\
\textbf{64}
& 70.33 $\pm$ 0.22 & 69.02 $\pm$ 0.19 & 93.60 $\pm$ 0.36
& 99.48 $\pm$ 0.07 & 97.10 $\pm$ 0.05 & 97.70 $\pm$ 0.22
& 99.55 $\pm$ 0.03 & 73.52 $\pm$ 0.44 & 87.54 $\pm$ 0.07 \\
128
& 70.66 $\pm$ 0.08 & 69.70 $\pm$ 0.24 & 94.09 $\pm$ 0.19
& 99.57 $\pm$ 0.04 & 97.22 $\pm$ 0.04 & 98.18 $\pm$ 0.18
& 99.64 $\pm$ 0.02 & 75.46 $\pm$ 0.21 & 88.07 $\pm$ 0.06 \\
256
& \textbf{70.76} $\pm$ 0.18 & \textbf{70.11} $\pm$ 0.27 & \textbf{94.33} $\pm$ 0.15
& \textbf{99.64} $\pm$ 0.03 & \textbf{97.29} $\pm$ 0.03 & \textbf{98.37} $\pm$ 0.02
& \textbf{99.67} $\pm$ 0.01 & \textbf{76.19} $\pm$ 0.34 & \textbf{88.29} $\pm$ 0.07 \\
\bottomrule
\end{tabular}
}
\end{table}

\end{document}